\documentclass[11pt]{article}

\usepackage[final]{acl}

\usepackage{times}
\usepackage{latexsym}

\usepackage[T1]{fontenc}
\usepackage[utf8]{inputenc}

\usepackage{microtype}

\usepackage{inconsolata}

\usepackage{graphicx}

\usepackage{amsmath}
\usepackage{amssymb}
\usepackage{amsfonts}

\usepackage{booktabs}
\usepackage{multirow}
\usepackage{caption}
\usepackage{array}
\usepackage[table]{xcolor}

\usepackage{enumitem}

\usepackage{algorithm}
\usepackage{algpseudocode}

\title{A Stronger Teacher Within: Strengthening MoE Models\\ via On-Policy Bidirectional Distillation}

\author{
Haoyu Chen$^{1}$, 
Jun Bai$^{3}$\thanks{Corresponding author.}, 
Yang Liu$^{3}$, 
Jianfei Zhang$^{4}$, 
Wenge Rong$^{2}$
\\[0.5ex]
\textsuperscript{1}National College for Excellent Engineers, Beihang University \\
\textsuperscript{2}School of Computer Science and Engineering, Beihang University \\
\textsuperscript{3}Beijing Institute for General Artificial Intelligence, BIGAI\\
\textsuperscript{4} Meituan Inc.\\%\\
}

\begin{document}
\maketitle
\begin{abstract}
Reinforcement Learning with Verifiable Rewards (RLVR) has substantially improved Mixture-of-Experts (MoE) LLMs, but its outcome-level rewards provide only sparse supervision. On-Policy Distillation offers dense token-level feedback, yet requires a stronger compatible teacher, while on-policy self-distillation relies on hand-designed privileged information. We observe that the MoE architecture itself offers a way to resolve this dilemma, as increasing the activation budget per token exposes latent expert capacity and yields a potentially stronger teacher within the model. However, naive expansion degrades performance because newly co-activated experts are not optimized to work together. We therefore propose MoEvolve, a bidirectional distillation framework following a recovery-then-improvement strategy: we first restore the activation-expanded high-$k$ model and further strengthen it by exploiting its additional active capacity, then reverse the direction of distillation to transfer the enhanced capabilities back into the original low-$k$ configuration, retaining its inference-time activation budget. Across diverse MoE models and benchmarks, MoEvolve improves the initial models by up to $22.2$ domain-average points and outperforms same-budget RLVR baselines by up to $6.5$ points. Notably, the final low-$k$ models often surpass their activation-expanded teachers.
% \footnote{Code is available at \url{https://anonymous.4open.science/r/moevolve}.}
\end{abstract}

\section{Introduction}

Mixture-of-Experts (MoE) has become a prominent architecture for state-of-the-art Large Language Models (LLMs) \citep[\textit{e.g.},][]{kimi2026k3,glm5}. By activating only a small subset of experts per token, MoE models offer a favorable capability-efficiency trade-off \citep{shazeer2017outrageously,fedus2022switch,deepseekai2026deepseekv4highlyefficientmilliontoken}.

Post-training methods such as Reinforcement Learning with Verifiable Rewards (RLVR) have improved MoE LLMs on reasoning tasks \citep{shao2024deepseekmath,guo2025deepseek,coreteam2026mimov2flashtechnicalreport}, but outcome-level rewards provide sparse supervision for intermediate steps. 
Process-level feedback provides a natural source of denser supervision, but training Process Reward Models (PRMs) requires costly step-level annotations \citep{pmlr}. 

Recently, On-Policy Distillation (OPD) has emerged as a promising alternative \citep{opd}. By supervising trajectories sampled from the student's current policy with token-level teacher logits, OPD provides dense feedback without requiring a specifically trained PRM \citep{yang2026learning,hou2026uniopd}.
However, OPD depends on a strong teacher model sharing compatible tokenizers and output spaces with the student model \citep{opd}, which may not always be available or affordable in practice. 
% Moreover, token-level supervision is most straightforward when the teacher and student share compatible tokenizers and output spaces. Otherwise, additional alignment or retokenization is required, increasing implementation complexity and potentially introducing approximation errors \citep{}.
On-policy self-distillation (OPSD) sidesteps this by using the student model itself as teacher with \emph{privileged information} (\textit{e.g.}, partial solutions, hints) \citep{opsd}, but this shifts the burden to designing privileged signals: they may require unavailable ground-truth solutions, and can encourage shortcut memorization rather than transferable capability learning, thereby hurting generalization \citep{zhao2026rosd,shen2026purified}.

In this paper, we observe that MoE LLMs offer a natural way out of this dilemma. A typical MoE routes each token to only $k$ of its $N$ experts, leaving substantial expert capacity inactive. Increasing the activation budget to $k' > k$ exposes each token to additional expert transformations and computation \citep{geva-etal-2021-transformer,dai-etal-2022-knowledge,huang2024harder}, creating an activation-expanded variant with the potential to serve as a stronger, fully compatible teacher. However, naively increasing $k$ in a pretrained MoE can trigger \emph{mode collapse}, as newly co-activated experts are not optimized to work together \citep{gu2025elastic,wang2025matryoshka}. 
We find that this degradation can be effectively repaired by applying OPD from the original low-$k$ model, whose dense token-level supervision restores the behavior disrupted by activation expansion. However, OPD alone is bounded by the capability of its teacher and offers limited incentive to exploit the additional active capacity. We therefore follow it with RLVR, which further strengthens the recovered high-$k$ model beyond the original one.

We then reverse the direction of distillation, using the strengthened high-$k$ model as the teacher to improve the original low-$k$ model. This transfers the gains from activation expansion back into the original inference configuration, preserving its activation budget and inference-time computation. We term the resulting bidirectional distillation framework \textbf{MoEvolve}. It enables MoE self-improvement without requiring either a stronger external teacher, as in OPD, or hand-designed privileged information, as in OPSD.

Our contributions are summarized as follows:
\begin{itemize}[leftmargin=1.5em]
    \item We propose MoEvolve, a self-improvement framework for MoE LLMs that exploits latent expert capacity to construct a stronger, fully compatible teacher within the model itself, and then distills the acquired capabilities back into the original low-$k$ configuration without increasing inference-time computation.
    \item We introduce a recovery-then-improvement strategy for constructing a stronger activation-expanded teacher: first restoring the expert coordination disrupted by activation expansion, and then exploiting the additional active capacity to improve the recovered model beyond its original capability.
    \item Extensive experiments across four MoE LLMs and twelve benchmarks spanning mathematics, code, and scientific knowledge show consistent gains at unchanged inference cost, with improvements of up to $22.2$ domain-average points over the initial models and up to $6.5$ points over strong same-budget RLVR baselines. Notably, the final low-$k$ models often surpass their activation-expanded teachers, suggesting that reverse distillation can produce genuine capability gains beyond lossy compression.
\end{itemize}

\section{Related Work}

\paragraph{Mixture-of-Experts (MoE) LLMs.}
MoE replaces the dense feed-forward network in each Transformer layer with expert FFNs and a router that activates only the top-$k$ experts per token \citep{shazeer2017outrageously,fedus2022switch}. Prior work studies expert specialization and load balancing \citep{huang2024harder,olmoe}. Distillation has also been used to compress MoE teachers into smaller models \citep{kim2025everyexpert,kim2026pruning} and to reduce expert activation through self-distillation \citep{lv2026zeda}. For this paper, the key observation is that simply raising the activation budget $k$ can hurt performance because newly co-activated experts were never optimized to cooperate \citep{gu2025elastic,wang2025matryoshka}. This motivates a way to restore coordination after expansion.

\paragraph{On-Policy Distillation (OPD).}
Knowledge distillation transfers the behavior of a teacher model into a student \citep{hinton2015distilling,sanh2019distilbert}. OPD trains the student on trajectories sampled from itself, which reduces exposure bias and provides dense token-level supervision through the teacher's logits \citep{opd,yang2026learning}. ExOPD further adds reward extrapolation so that the student can benefit more fully from the stronger teacher \citep{yang2026learning}.
However, OPD relies on a strong teacher with a compatible tokenizer and output space. Such a teacher may not always be available or affordable in practice \citep{opd}. To avoid this external dependency, OPSD leverages privileged information to act as its own teacher, distilling a model to itself without increasing inference cost \citep{opsd,zhao2026rosd,shen2026purified}. Such privileged signals often include partial solutions, hints, or reference answers \citep{opsd}. Designing these signals can be task-specific, may require unavailable ground-truth solutions, and can hurt generalization by encouraging imitation of reference trajectories \citep{zhao2026rosd,shen2026purified}.
% In our setting, the model's \emph{own} activation-expanded variant serves as the teacher, and the recovered capability is distilled back into the \emph{same} low-$k$ model while keeping its activation budget unchanged.

% \begin{figure*}[t]
%   \centering
%   \includegraphics[width=1.0\textwidth]{pics/method-fancy_pic_ong.pdf}
%   \caption{Overview of MoEvolve. We increase the activation budget from $k$ to $k'$, restore expert cooperation (ECRD), further strengthen the activation-expanded model with RL (CU), and finally use it as a teacher to improve the original $k$-activation model without changing its inference-time budget (RD).}
%   \label{fig:method}
% \end{figure*}

\begin{figure*}[t]
  \centering
  \includegraphics[width=1.0\textwidth]{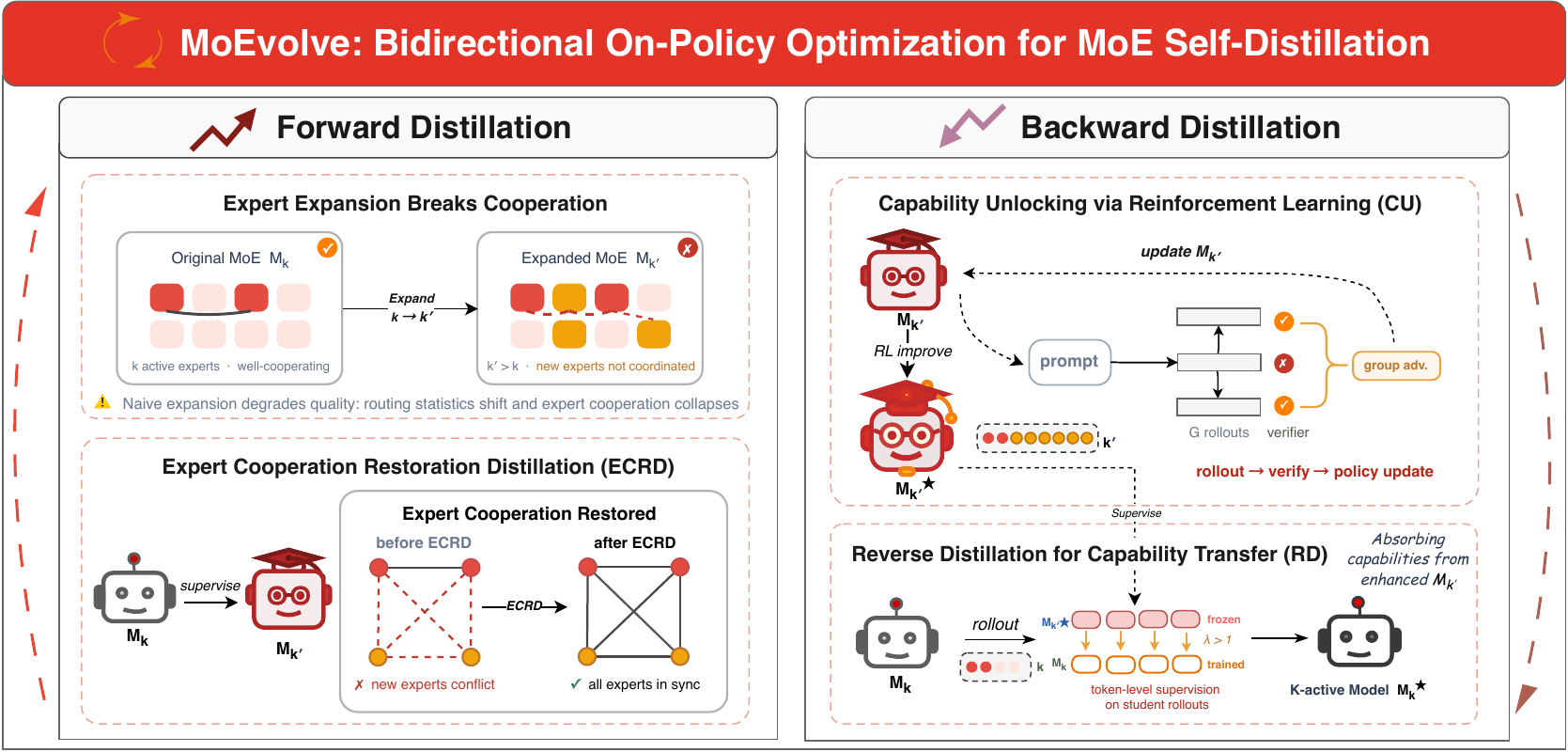}
  \caption{Overview of MoEvolve. We increase the activation budget from $k$ to $k'$, restore expert cooperation (ECRD), further strengthen the activation-expanded model with RL (CU), and finally use it as a teacher to improve the original $k$-activation model without changing its inference-time budget (RD).}
  \label{fig:method}
\end{figure*}

\section{Preliminaries}
\label{sec:prelim}

We review the three components underlying our method. The MoE formulation defines the activation budget, whose expansion from $k$ to $k'$ enables us to construct an internal teacher. OPD provides the objectives for restoring cooperation in the expanded model and transferring its gains back to the original budget. SAPO \citep{sapo} is used to improve the MoE model while mitigating training instability caused by mismatch between training and rollout.

\paragraph{Mixture-of-Experts.} Given a hidden representation $\mathbf{h} \in \mathbb{R}^d$, the router produces affinity scores
\begin{equation}
    g_i(\mathbf{h}) = \frac{\exp(\mathbf{w}_i^\top \mathbf{h})}{\sum_{j=1}^{N} \exp(\mathbf{w}_j^\top \mathbf{h})},
    \label{eq:router}
\end{equation}
where $\mathbf{w}_i$ is the routing vector of expert $i$, and $N$ is the total number of experts. Let $\mathcal{T}_k(\mathbf{h})$ denote the top-$k$ experts selected by the router. The MoE layer outputs $\mathbf{y} = \sum_{i \in \mathcal{T}_k(\mathbf{h})} \tilde{g}_i(\mathbf{h}) \, E_i(\mathbf{h})$, where $\tilde{g}_i(\mathbf{h}) = g_i(\mathbf{h}) / \sum_{j \in \mathcal{T}_k(\mathbf{h})} g_j(\mathbf{h})$ are normalized routing weights. Since only $k \ll N$ experts are activated per token, cost depends on $k$ rather than $N$, allowing capacity to scale without proportional compute. We denote the original model by $\mathcal{M}_k$ and the activation-expanded variant by $\mathcal{M}_{k'}$. $\mathcal{M}_k$ and $\mathcal{M}_{k'}$ are two separate model copies with identical initial expert and routing parameters. They differ only in the activation budget, namely $k$ versus $k'$.

\paragraph{On-Policy Distillation.} Given a prompt $\mathbf{x}$, the student samples a response $\mathbf{y} \sim \pi_S(\cdot \mid \mathbf{x})$, and OPD minimizes the reverse KL divergence between the teacher and student on that response
\begin{equation}
\begin{split}
&\mathcal{J}_{\text{OPD}}(\pi_T,\pi_S) = \min_{\theta_S}\; \mathbb{E}_{\mathbf{x} \sim \mathcal{D},\, \mathbf{y} \sim \pi_S} \Big[ \\
&\qquad D_{\mathrm{KL}}\big( \pi_S(\mathbf{y} \mid \mathbf{x}) \,\| \mathrm{sg}\bigl[\pi_T(\mathbf{y} \mid \mathbf{x})\bigr] \big) \Big].
\end{split}
\label{eq:opd}
\end{equation}
Here the KL is at the token level over next token distributions along the student-sampled prefix, $\mathrm{sg}[\cdot]$ denotes stop-gradient on the teacher logits, and $\mathcal{D}$ is the prompt distribution. Standard OPD fixes the teacher signal and the KL regularization to equal weight. Generalized On-Policy Distillation (G-OPD) generalizes this view by adding a reference policy $\pi_{\mathrm{ref}}$ and a reward scaling factor $\lambda$ \citep{yang2026learning}:
\begin{equation}
\begin{split}
&\mathcal{J}_{\mathrm{G\text{-}OPD}}(\pi_S,\pi_T,\pi_{\mathrm{ref}},\lambda) \\
&\quad = \max_{\theta_S}\; \mathbb{E}_{\mathbf{x}\sim\mathcal{D},\,\mathbf{y}\sim\pi_S}
\Bigg[ \lambda \log \frac{\pi_T(\mathbf{y}\mid\mathbf{x})}{\pi_{\mathrm{ref}}(\mathbf{y}\mid\mathbf{x})} \\
&\qquad \quad - D_{\mathrm{KL}}\!\left( \pi_S(\mathbf{y}\mid\mathbf{x}) \,\|\, \pi_{\mathrm{ref}}(\mathbf{y}\mid\mathbf{x}) \right) \Bigg].
\end{split}
\label{eq:gopd}
\end{equation}
Standard OPD is the special case $\lambda=1$. G-OPD uses $\lambda>1$ for reward extrapolation, termed ExOPD, giving extra weight to a strong teacher. Prior work demonstrates that ExOPD significantly outperforms standard OPD when distilling between models of the same family \citep{yang2026learning}. In our setting, the teacher and student are naturally paired copies of the same MoE models with different activation budgets, so we use ExOPD in the reverse-distillation stage.

\paragraph{Soft Adaptive Policy Optimization (SAPO).} SAPO \citep{sapo} replaces hard clipping in GRPO \citep{guo2025deepseek} with a smooth temperature-controlled gate. For $G$ responses sampled from the behavior policy, it optimizes
\begin{equation}
\begin{aligned}
&\mathcal{J}_{\mathrm{SAPO}}(\theta) = \max_{\theta}\; \mathbb{E}_{\mathbf{x}\sim\mathcal{D},\,\mathbf{y}\sim\pi_\theta(\cdot\mid\mathbf{x})}\Bigg[ \\
&\quad \sum_{i=1}^{G}\frac{1}{G|\mathbf{y}_i|}
\sum_{t=1}^{|\mathbf{y}_i|} f^{\mathrm{SAPO}}_{i,t}(r_{i,t})\hat{A}_{i,t}\Bigg],
\end{aligned}
\label{eq:stage2}
\end{equation}
where $r_{i,t}$ is the token-level importance ratio, $\hat{A}_{i,t}$ is the group relative advantage, and $f^{\mathrm{SAPO}}_{i,t}(r)=\frac{4}{\tau_i}\sigma(\tau_i(r-1))$ with $\tau_i=\tau_{\mathrm{pos}}$ for positive advantages and $\tau_i=\tau_{\mathrm{neg}}$ otherwise. We set $\tau_{\mathrm{neg}}>\tau_{\mathrm{pos}}$ to downweight high-variance negative updates, yielding a smooth trust region that stabilizes RL on MoE models.

\section{MoEvolve: Bidirectional On-Policy Optimization for MoE Self-Distillation}
% \section{Self-Forge: Building a Stronger MoE Teacher from Itself}
\label{sec:method}

We now present MoEvolve, our activation-expansion based training framework, illustrated in Figure~\ref{fig:method}. Starting from a trained MoE model $\mathcal{M}_k$, the framework proceeds in two phases. \textbf{Forward Distillation} expands the activation budget from $k$ to $k'$ and runs Expert Cooperation Restoration Distillation (ECRD), which takes $\mathcal{M}_k$ as the teacher to restore coordination among the newly co-activated experts. \textbf{Backward Distillation} then flips the direction of supervision. It first applies Capability Unlocking via Reinforcement Learning (CU) to push the recovered activation-expanded model beyond $\mathcal{M}_k$ under the larger budget, yielding a stronger internal teacher $\mathcal{M}_{k'}^{\star}$, and then performs Reverse Distillation for Capability Transfer (RD), which distills that unlocked capability back into the original $k$-expert budget. Together, these two phases produce an improved $k$-activation model $\mathcal{M}_k^{\star}$ with unchanged inference cost.

\subsection{Forward Distillation}
\label{sec:method-stage1}

Increasing the number of activated experts widens the active intermediate representation of each MoE layer (Appendix~\ref{sec:appendix-capacity}) and exposes each token to additional expert transformations and computation, providing additional representational capacity. However, directly raising the activation budget of a trained model from $k$ to $k'$ yields an activation-expanded model $\mathcal{M}_{k'}$ whose co-activated experts were never trained to work together. As a result, $\mathcal{M}_{k'}$ suffers mode collapse and can perform \emph{worse} than $\mathcal{M}_k$ despite using more FFN capacity \citep{gu2025elastic,wang2025matryoshka}. Our first step therefore restores expert cooperation, corresponding to the ECRD stage in Figure~\ref{fig:method}.

Many recovery choices are possible at this stage, including supervised fine-tuning (SFT) \citep{guha2025openthoughts,team2025kimi}, rejection sampling fine-tuning (RFT) \citep{shao2024deepseekmath}, reinforcement learning (RL), and OPD. In our method, the ECRD is implemented with OPD: the original model $\mathcal{M}_k$ serves as the teacher and the activation-expanded model $\mathcal{M}_{k'}$ serves as the student. We choose this ECRD design because SFT requires reference solutions from an external source, which does not fit our goal of self-improvement without an external teacher. RFT can use verified samples from $\mathcal{M}_k$, but it still trains on a fixed off-policy set and is limited by the behaviors captured in that set. Direct RL avoids such targets, yet outcome-only rewards provide sparse sequence-level feedback. With uncoordinated experts, this easily leads to entropy collapse and stagnant reward, as shown in Appendix~\ref{sec:appendix-reward-entropy}. By contrast, ECRD samples on-policy from $\mathcal{M}_{k'}$ and uses $\mathcal{M}_k$ as an internal teacher, providing dense supervision without external labels.

Concretely, in ECRD we optimize the objective in Eq.~\eqref{eq:opd} with $\pi_T = \pi_{\mathcal{M}_k}$ and $\pi_S = \pi_{\mathcal{M}_{k'}}$. The trajectories are sampled from the student $\mathcal{M}_{k'}$ itself, so the teacher supervises exactly the states the activation-expanded model actually visits. This quickly pulls the newly co-activated experts back into a cooperative regime. In practice this step rapidly recovers the performance lost to mode collapse, so that $\mathcal{M}_{k'}$ at least matches $\mathcal{M}_k$ while now genuinely leveraging its larger activation budget.

\subsection{Backward Distillation}

\subsubsection{Capability Unlocking via Reinforcement Learning (CU)}
\label{sec:method-stage2}

ECRD restores cooperation, but it has yet to harness the stronger representational capacity brought by activating more experts to improve the model’s capability. To unlock this capacity and obtain a stronger teacher, we continue training $\mathcal{M}_{k'}$ with reinforcement learning using SAPO \citep{sapo}, corresponding to the CU stage in Figure~\ref{fig:method}. Because the experts already cooperate after ECRD, RL optimization becomes stable and effective. This step allows $\mathcal{M}_{k'}$ to realize the benefit of the larger activation budget and \emph{surpass} the original model $\mathcal{M}_k$, at the temporary cost of activating $k'$ experts. We denote the resulting strengthened activation-expanded model by $\mathcal{M}_{k'}^{\star}$.

% \subsection{Reverse Distillation into the Original Model}
\subsubsection{Reverse Distillation for Capability Transfer (RD)}
\label{sec:method-stage3}

After the CU step we obtain a stronger model $\mathcal{M}_{k'}^{\star}$, but it activates $k' > k$ experts and therefore remains more expensive at inference time. The final step transfers this gain back to the original activation budget by \emph{reversing} the distillation direction. We use ExOPD, instantiated by the objective in Eq.~\eqref{eq:gopd}, with $\pi_S = \pi_{\mathcal{M}_k}$, $\pi_T = \pi_{\mathcal{M}_{k'}^{\star}}$, $\pi_{\mathrm{ref}} = \pi_{\mathcal{M}_k}$, and $\lambda>1$. As discussed in \S\ref{sec:prelim}, ExOPD has shown stronger distillation performance than OPD on the same family of models, motivating its use here. This fits our paired-model setting: $\mathcal{M}_{k'}^{\star}$ and $\mathcal{M}_k$ share the tokenizer, vocabulary, and architecture, and mainly differ in the capability recovered through the larger activation budget. The teacher-reference log-ratio in Eq.~\eqref{eq:gopd} can therefore provide a clean dense signal for amplifying the useful changes in $\mathcal{M}_{k'}^{\star}$ while anchoring the student to the original $k$-activation model. The final model $\mathcal{M}_k^{\star}$ keeps the original $k$ experts per token, preserving the inference speed and activation cost.

\section{Experiments}
\label{sec:experiments}

\subsection{Experimental Setup}
\label{sec:exp-setup}

\paragraph{Models} We evaluate our method on four widely used open-source MoE LLMs spanning different model scales and expert configurations: OLMoE-1B-7B \citep{olmoe}, DeepSeek-V2-Lite \citep{deepseekv2}, GPT-OSS-20B \citep{gptoss}, and Qwen3-30B-A3B \citep{qwen2025qwen3}. These models cover a diverse range of MoE architectures and activation patterns, enabling us to assess whether our method generalizes across different model scales and expert configurations. Detailed model configurations are provided in Appendix~\ref{sec:appendix-models}.

\paragraph{Evaluation} We evaluate on three complementary domains to comprehensively assess the model's capabilities. \textbf{Mathematics} measures the model's capacity for symbolic reasoning, \textbf{Code generation} reflects its mastery of procedural logic and \textbf{Scientific and general knowledge} (\textbf{ScienceQA}) gauges the model's broad knowledge retention and application. Together, these diverse domains show that the improvements from MoEvolve are not limited to a specific task or problem setting. For \textbf{mathematics} we use AMC23 \citep{amc2023} together with GSM8K \citep{cobbe2021training}, MATH500 \citep{hendrycks2021math}, Minerva Math \citep{lewkowycz2022solving}, and PolyMath-Low (Poly-L) \citep{polymath} for OLMoE-1B-7B and DeepSeek-V2-Lite, and the harder PolyMath-Medium (Poly-M) \citep{polymath} and AIME-2024/2025/2026 \citep{aime2024,aime2025,aime2026} for GPT-OSS-20B and Qwen3-30B-A3B, matching benchmark difficulty to backbone strength. For \textbf{code generation} we use HumanEval (HEval) \citep{chen2021evaluating}, MBPP (full and MBPP-Sanitized) \citep{austin2021program}, and LiveCodeBench-v6 (LCB) \citep{jain2024livecodebench}. For \textbf{ScienceQA} we use GPQA-main \citep{rein2024gpqa}, ARC-Challenge \citep{clark2018think}, and MMLU-STEM \citep{hendrycks2021measuring}. We report pass@1 \citep{chen2021evaluating} across all benchmarks, sampling $10$ generations per problem. Detailed evaluation settings are provided in Appendix~\ref{sec:appendix-eval}.

\paragraph{Training Datasets and Settings} We train on three domains, matching the three evaluation domains. For mathematics we use OpenMathInstruct-2 \citep{openmathinstruct2} and DeepMath-103K \citep{deepmath}, for code generation we use the code split of Eurus-2-RL-Data \citep{eurusrlcode}, and for ScienceQA we use the MMLU auxiliary training set \citep{hendrycks2021measuring}. Detailed training settings are in Appendix~\ref{sec:appendix-impl}.

\paragraph{Baselines} We compare against representative methods that improve $\mathcal{M}_k$ \emph{without} raising its activation budget: (1) \textbf{SFT}, supervised fine-tuning of $\mathcal{M}_k$ on the reference solutions of the mathematics corpus, i.e.\ the Llama-3.1-405B-Instruct solutions of OpenMathInstruct-2 \citep{openmathinstruct2} and the DeepSeek-R1 traces of DeepMath-103K \citep{deepmath}, the only domain shipping such labels, hence run on mathematics only \citep{guha2025openthoughts,team2025kimi}. (2) \textbf{RFT}, rejection fine-tuning of $\mathcal{M}_k$ using its own verifier-verified responses as labels \citep{shao2024deepseekmath}. (3) \textbf{RL}, reinforcement learning directly on $\mathcal{M}_k$ using SAPO \citep{shao2024deepseekmath,guo2025deepseek}. (4) \textbf{OPSD}, on-policy self-distillation where the teacher is $\mathcal{M}_k$ itself prompted with privileged information \citep{opsd,zhao2026rosd,shen2026purified,sdpo}. Detailed privileged information settings are given in Appendix~\ref{sec:appendix-training}.
All baselines and our method start from the same checkpoint.

\begin{table*}[t]
  \centering
  \footnotesize
  \setlength{\tabcolsep}{1.6pt}
  \resizebox{\textwidth}{!}{%
  \begin{tabular}{lccc>{\columncolor{gray!12}}c ccccc>{\columncolor{gray!12}}c cccc>{\columncolor{gray!12}}c}
    \toprule
    \multirow{2}{*}{\textbf{Method}} & \multicolumn{4}{c}{\textbf{ScienceQA}} & \multicolumn{6}{c}{\textbf{Mathematics}} & \multicolumn{5}{c}{\textbf{Code generation}} \\
    \cmidrule(lr){2-5} \cmidrule(lr){6-11} \cmidrule(lr){12-16}
    & \textbf{GPQA} & \textbf{ARC-C} & \textbf{MMLU-S} & \textbf{Avg.} & \textbf{GSM8K} & \textbf{MATH500} & \textbf{Minerva} & \textbf{AMC23} & \textbf{Poly-L} & \textbf{Avg.} & \textbf{HEval} & \textbf{MBPP} & \textbf{MBPP-S} & \textbf{LCB} & \textbf{Avg.} \\
    \midrule
    \multicolumn{16}{l}{\textbf{\emph{OLMoE-1B-7B}}} \\
    Original $\mathcal{M}_k$   & 21.5 & 46.6 & 32.1 & 33.4$_{\pm 0.12}$ & 27.5 & 7.1 & 2.1 & 5.5 & 25.6 & 13.6$_{\pm 0.08}$ & 25.0 & 19.2 & 30.0 & 1.1 & 18.8$_{\pm 0.09}$ \\
    SFT                        & -- & -- & -- & -- & 30.4 & 7.5 & 1.9 & 5.5 & 30.0 & 15.1$_{\pm 0.11}$ & -- & -- & -- & -- & -- \\
    RFT                        & 26.2 & 59.4 & 42.5 & 42.7$_{\pm 0.13}$ & 34.2 & 7.8 & 2.8 & 5.0 & 30.6 & 16.1$_{\pm 0.09}$ & 27.2 & 20.3 & 31.8 & 2.4 & 20.4$_{\pm 0.10}$ \\
    RL                         & 26.5 & 63.3 & 43.9 & 44.6$_{\pm 0.15}$ & 34.6 & 7.9 & 2.8 & 4.8 & 30.9 & 16.2$_{\pm 0.10}$ & 27.4 & 20.5 & 32.0 & 2.3 & 20.6$_{\pm 0.12}$ \\
    % RL\,+\,OPD                 & 25.9 & 61.4 & 42.2 & 43.2 & 37.3 & 7.5 & 2.7 & 5.0 & 33.8 & 17.3 & 25.3 & 21.4 & 32.5 & 1.2 & 20.1 \\
    OPSD                       & 22.8 & 54.9 & 36.6 & 38.1$_{\pm 0.14}$ & 27.9 & 7.4 & 2.1 & \textbf{5.8} & 27.8 & 14.2$_{\pm 0.09}$ & 25.8 & 19.9 & 30.8 & 1.6 & 19.5$_{\pm 0.11}$ \\
    \textbf{Ours}  & \textbf{27.9} & \textbf{64.0} & \textbf{45.2} & \textbf{45.7}$_{\pm 0.14}$ & \textbf{36.2} & \textbf{9.3} & \textbf{3.7} & \textbf{5.8} & \textbf{33.4} & \textbf{17.7}$_{\pm 0.10}$ & \textbf{30.0} & \textbf{21.7} & \textbf{34.4} & \textbf{4.1} & \textbf{22.6}$_{\pm 0.12}$ \\
    \midrule
    \multicolumn{16}{l}{\textbf{\emph{DeepSeek-V2-Lite}}} \\
    Original $\mathcal{M}_k$   & 21.3 & 71.6 & 49.6 & 47.5$_{\pm 0.13}$ & 63.2 & 22.3 & 8.1 & 7.0 & 61.9 & 32.5$_{\pm 0.10}$ & 47.1 & 30.9 & 42.7 & 7.8 & 32.1$_{\pm 0.11}$ \\
    SFT                        & -- & -- & -- & -- & 66.8 & 22.8 & 9.2 & 7.5 & 65.8 & 34.4$_{\pm 0.13}$ & -- & -- & -- & -- & -- \\
    RFT                        & 27.2 & 75.3 & 55.0 & 52.5$_{\pm 0.14}$ & 66.0 & 23.1 & 9.0 & 8.3 & 65.0 & 34.3$_{\pm 0.11}$ & 47.7 & 32.1 & 44.0 & 9.4 & 33.3$_{\pm 0.12}$ \\
    RL                         & 27.9 & \textbf{76.2} & 55.4 & 53.2$_{\pm 0.14}$ & 66.8 & 23.2 & 9.0 & 8.5 & 66.8 & 34.9$_{\pm 0.11}$ & 47.8 & 32.2 & 44.0 & 9.4 & 33.4$_{\pm 0.12}$ \\
    OPSD                       & 25.9 & 73.4 & 54.9 & 51.4$_{\pm 0.15}$ & 63.7 & 22.7 & 7.5 & 7.8 & 64.9 & 33.3$_{\pm 0.12}$ & 47.6 & 31.6 & 43.5 & 8.6 & 32.8$_{\pm 0.13}$ \\
    % RL\,+\,OPD                 & 27.1 & 76.0 & 54.7 & 52.6 & 65.0 & 22.9 & 8.5 & 5.0 & 63.3 & 32.9 & 47.6 & 31.0 & 42.3 & 9.2 & 32.5 \\
    \textbf{Ours}  & \textbf{28.9} & 76.1 & \textbf{56.4} & \textbf{53.8}$_{\pm 0.13}$ & \textbf{68.2} & \textbf{24.2} & \textbf{9.8} & \textbf{10.3} & \textbf{68.6} & \textbf{36.2}$_{\pm 0.10}$ & \textbf{49.4} & \textbf{35.8} & \textbf{48.5} & \textbf{9.6} & \textbf{35.8}$_{\pm 0.11}$ \\
    \bottomrule
  \end{tabular}}
  \caption{Main results on OLMoE-1B-7B and DeepSeek-V2-Lite (pass@1, \%). \textbf{Avg.} denotes the average score within each domain. SFT covers only mathematics, so its missing entries are marked ``--''. Error bars indicate the standard error across 3 seeds.}
  \label{tab:main-ds-olmoe}
\end{table*}

\begin{table*}[t]
  \centering
  \footnotesize
  \setlength{\tabcolsep}{1.6pt}
  \resizebox{\textwidth}{!}{%
  \begin{tabular}{lccc>{\columncolor{gray!12}}c ccccc>{\columncolor{gray!12}}c cccc>{\columncolor{gray!12}}c}
    \toprule
    \multirow{2}{*}{\textbf{Method}} & \multicolumn{4}{c}{\textbf{ScienceQA}} & \multicolumn{6}{c}{\textbf{Mathematics}} & \multicolumn{5}{c}{\textbf{Code generation}} \\
    \cmidrule(lr){2-5} \cmidrule(lr){6-11} \cmidrule(lr){12-16}
    & \textbf{GPQA} & \textbf{ARC-C} & \textbf{MMLU-S} & \textbf{Avg.} & \textbf{AMC23} & \textbf{Poly-M} & \textbf{AIME24} & \textbf{AIME25} & \textbf{AIME26} & \textbf{Avg.} & \textbf{HEval} & \textbf{MBPP} & \textbf{MBPP-S} & \textbf{LCB} & \textbf{Avg.} \\
    \midrule
    \multicolumn{16}{l}{\textbf{\emph{GPT-OSS-20B}}} \\
    Original $\mathcal{M}_k$  & 43.5 & 82.7 & 70.4 & 65.5$_{\pm 0.16}$ & 84.8 & 48.5 & 41.3 & 38.0 & 43.0 & 51.1$_{\pm 0.14}$ & 93.0 & 52.0 & 67.5 & 47.6 & 65.0$_{\pm 0.15}$ \\
    SFT                       & -- & -- & -- & -- & 85.8 & 51.0 & 45.3 & 40.0 & 46.3 & 53.7$_{\pm 0.17}$ & -- & -- & -- & -- & -- \\
    RFT                       & 48.6 & 91.5 & 79.4 & 73.2$_{\pm 0.18}$ & 86.5 & 52.5 & 46.7 & 39.7 & 47.7 & 54.6$_{\pm 0.17}$ & 93.4 & 54.7 & 70.1 & 56.6 & 68.7$_{\pm 0.18}$ \\
    RL                        & \textbf{50.0} & 91.1 & 80.4 & 73.8$_{\pm 0.13}$ & 86.8 & 52.3 & 47.0 & 41.7 & 48.3 & 55.2$_{\pm 0.12}$ & 93.5 & 54.8 & 70.3 & 57.5 & 69.0$_{\pm 0.13}$ \\
    % RL\,+\,OPD                & 48.4 & 90.3 & 79.9 & 72.9 & 84.5 & 49.2 & 45.0 & 39.3 & 45.0 & 52.6 & 93.2 & 53.3 & 67.9 & 56.7 & 67.8 \\
    OPSD                      & 46.2 & 84.4 & 73.8 & 68.1$_{\pm 0.18}$ & 85.5 & 49.0 & 41.7 & 39.0 & 41.3 & 51.3$_{\pm 0.16}$ & 92.8 & 52.7 & 68.9 & 51.2 & 66.4$_{\pm 0.17}$ \\
    \textbf{Ours} & 49.7 & \textbf{93.2} & \textbf{83.3} & \textbf{75.4}$_{\pm 0.14}$ & \textbf{90.5} & \textbf{56.0} & \textbf{51.3} & \textbf{45.7} & \textbf{53.0} & \textbf{59.3}$_{\pm 0.13}$ & \textbf{93.7} & \textbf{56.6} & \textbf{72.3} & \textbf{59.8} & \textbf{70.6}$_{\pm 0.14}$ \\
    \midrule
    \multicolumn{16}{l}{\textbf{\emph{Qwen3-30B-A3B}}} \\
    Original $\mathcal{M}_k$   & 43.7 & 95.4 & 85.1 & 74.7$_{\pm 0.15}$ & 71.5 & 48.3 & 27.0 & 20.3 & 20.7 & 37.6$_{\pm 0.13}$ & 91.5 & 56.5 & 72.0 & 28.6 & 62.1$_{\pm 0.14}$ \\
    SFT& -- & -- & -- & -- & 85.5 & 59.3 & 38.3 & 29.7 & 34.7 & 49.5$_{\pm 0.19}$ & -- & -- & -- & -- & -- \\
    RFT                        & 54.7 & 96.3 & 92.6 & 81.2$_{\pm 0.15}$ & 86.3 & 58.4 & 40.0 & 32.3 & 36.7 & 50.7$_{\pm 0.14}$ & 93.3 & 60.3 & 75.2 & 36.9 & 66.4$_{\pm 0.15}$ \\
    RL                         & 54.6 & 96.4 & 92.2 & 81.1$_{\pm 0.19}$ & 88.5 & 60.6 & 45.3 & 32.7 & 39.3 & 53.3$_{\pm 0.18}$ & 94.1 & 61.7 & 76.4 & 39.9 & 68.0$_{\pm 0.19}$ \\
    % RL\,+\,OPD                 & 52.2 & 96.1 & 90.9 & 79.7 & 87.0 & 56.8 & 44.7 & 32.3 & 35.7 & 51.3 & 92.2 & 60.8 & 74.9 & 37.3 & 66.3 \\
    OPSD                       & 48.6 & 95.6 & 86.2 & 76.8$_{\pm 0.16}$ & 73.8 & 55.2 & 33.3 & 28.7 & 31.3 & 44.5$_{\pm 0.15}$ & 92.4 & 58.5 & 73.1 & 35.3 & 64.8$_{\pm 0.16}$ \\
    \textbf{Ours} & \textbf{56.5} & \textbf{96.6} & \textbf{92.9} & \textbf{82.0}$_{\pm 0.18}$ & \textbf{90.3} & \textbf{65.4} & \textbf{54.0} & \textbf{42.0} & \textbf{47.3} & \textbf{59.8}$_{\pm 0.17}$ & \textbf{94.6} & \textbf{62.3} & \textbf{78.0} & \textbf{41.4} & \textbf{69.1}$_{\pm 0.18}$ \\
    \bottomrule
  \end{tabular}}
  \caption{Main results on GPT-OSS-20B and Qwen3-30B-A3B (pass@1, \%). \textbf{Avg.} denotes the average score within each domain. SFT covers only mathematics, so its missing entries are marked ``--''. Error bars indicate the standard error across 3 seeds.}
  \label{tab:main-gptoss}
\end{table*}

\subsection{Main Results}
\label{sec:exp-main}

Tables~\ref{tab:main-ds-olmoe} and~\ref{tab:main-gptoss} compare our training framework against the baselines across all four backbones. For \textbf{Ours}, we use the maximum feasible intermediate budget for each backbone ($k'=64$ for OLMoE-1B-7B and DeepSeek-V2-Lite, $k'=32$ for GPT-OSS-20B, and $k'=128$ for Qwen3-30B-A3B). Our method achieves the best domain average on every backbone and every domain. Over the original model it gains up to $+12.3$ on ScienceQA, $+22.2$ on mathematics, and $+7.0$ on code, and it outperforms the strongest fixed-budget baseline (RL) by up to $+1.6$, $+6.5$, and $+2.4$ points on the three domains respectively. To confirm the gains are not seed-to-seed noise, we apply a paired $t$-test to the domain-average scores of Ours versus RL across the three seeds. Ours is significantly better on every backbone and domain ($p<0.05$). This indicates that transferring capability from the recovered activation-expanded model back into the original activation budget is more effective than improving the model directly. Detailed training time and compute comparisons are provided in Appendix~\ref{sec:appendix-compute}. Our method attains the best MMLU-STEM score on every backbone and the best GPQA score on every backbone except GPT-OSS-20B, indicating that restoring and then re-internalizing expert cooperation improves knowledge-intensive reasoning as well as mathematics and code.

\paragraph{Training dynamics and expert load versus baselines.} Beyond benchmark performance, we also compare training dynamics and expert utilization. On Qwen3-30B-A3B, our framework finishes with the highest reward while preserving the most balanced expert usage among the strong baselines, staying closest to the inherited routing pattern of $\mathcal{M}_k$. Appendix~\ref{sec:appendix-analysis-dynamics} gives the detailed training and routing analysis.

\subsection{Expert Cooperation Restoration Methods for the Activation-Expanded Model}
\label{sec:exp-recovery}

\begin{figure}[t]
  \centering
  \includegraphics[width=\columnwidth]{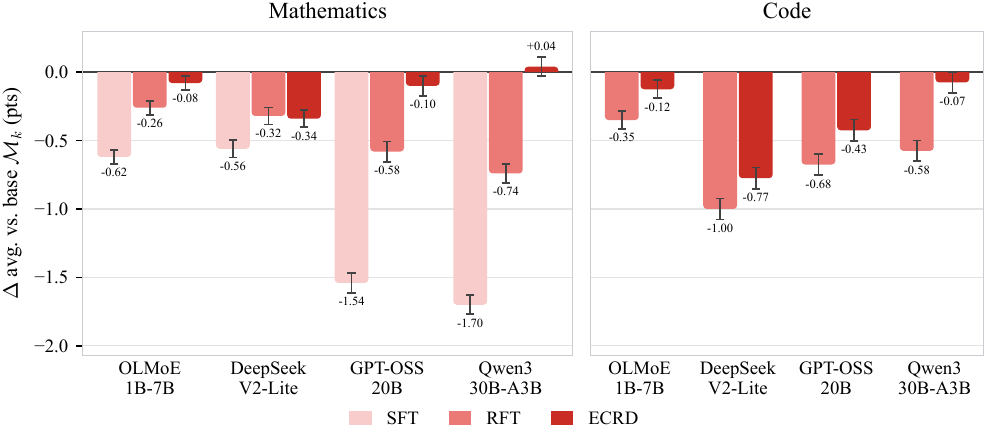}
  \caption{Recovery of the activation-expanded model $\mathcal{M}_{k'}$, shown as the gap of each recovery strategy relative to the original model $\mathcal{M}_k$ ($y{=}0$) on the mathematics average (left) and code average (right). Error bars show standard deviation across 3 seeds.}
  \label{fig:recovery}
\end{figure}

We next study how best to recover the activation-expanded model $\mathcal{M}_{k'}$ after its activation budget is expanded. Applying RL directly to $\mathcal{M}_{k'}$ diverges because the newly co-activated experts have not yet learned to cooperate, RL must optimize on top of a collapsed routing distribution and training collapses (Appendix~\ref{sec:appendix-reward-entropy}). Figure~\ref{fig:recovery} compares three stable recovery strategies. \textbf{SFT} fine-tunes $\mathcal{M}_{k'}$ on the reference solutions shipped with the training set. \textbf{RFT} first samples responses from the original model $\mathcal{M}_k$ on the same prompts, keeps those whose final answer is verified correct, and fine-tunes $\mathcal{M}_{k'}$ on them. \textbf{ECRD} takes $\mathcal{M}_k$ as the teacher and $\mathcal{M}_{k'}$ as the student and distills token-level supervision on the student's own trajectories (Eq.~\eqref{eq:opd}).

As shown in Figure~\ref{fig:recovery}, both SFT and RFT underperform ECRD. They are off-policy, since they fine-tune $\mathcal{M}_{k'}$ on a fixed set of target trajectories. SFT uses external reference solutions and RFT uses the original model's own verified responses, but in either case the training targets come from a policy different from the one $\mathcal{M}_{k'}$ samples from at inference. This causes exposure bias and errors accumulate once $\mathcal{M}_{k'}$ generates outside the imitation set. As an on-policy method, ECRD samples trajectories directly from $\mathcal{M}_{k'}$ and supervises them using the teacher's token-level logits. By providing supervision on the states visited by the student, ECRD aligns the training distribution with the student's inference distribution. This is why ECRD restores expert cooperation more faithfully than the off-policy alternatives.

\subsection{Capability Unlocked by the CU Stage}
\label{sec:exp-cu}

\begin{figure}[t]
  \centering
  \includegraphics[width=1.0\columnwidth]{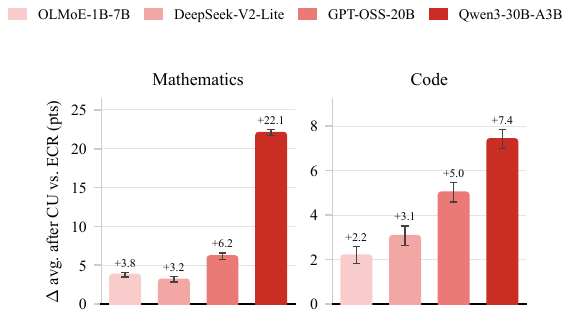}
  \caption{Capability unlocked by the CU stage, measured as the gain from the recovered activation-expanded model after ECRD to the same model after CU. Error bars show standard deviation across 3 seeds.}
  \label{fig:cu-unlock}
\end{figure}

Figure~\ref{fig:cu-unlock} quantifies the gain from the recovered $\mathcal{M}_{k'}$ after ECRD to the same model after CU. CU yields consistent improvements on all four backbones, ranging from $+3.2$ to $+22.1$ points on mathematics and from $+2.2$ to $+7.4$ points on code. The effect is especially large on Qwen3-30B-A3B, where the recovered model still leaves the most headroom after expansion. This shows that ECRD mainly repairs expert cooperation, whereas CU is the step that actually unlocks the additional active capacity and turns $\mathcal{M}_{k'}$ into a stronger teacher.

\subsection{How Best to Transfer Capability in RD}
\label{sec:exp-reverse}

We compare three objectives for transferring the capability of $\mathcal{M}_{k'}^{\star}$ back to $\mathcal{M}_k$: \textbf{RFT}, \textbf{OPD}, and \textbf{ExOPD}. The ranking ExOPD\,$>$\,OPD\,$>$\,RFT is consistent across backbones, with reward extrapolation contributing the largest gain. Detailed results and analysis are provided in Appendix~\ref{sec:appendix-rd-ablation}.

\subsection{Effect of the Intermediate Activation Budget}
\label{sec:exp-kprime}

We study how the intermediate activation budget $k'$ affects the final reverse-distilled model. We find that activating the largest feasible number of experts generally gives the best final result, so the method does not require careful tuning of $k'$. See Appendix~\ref{sec:exp-numexperts} for detailed analysis.

\section{Analyses}
\label{sec:analysis}

\subsection{Analysis of Expert Cooperation after Activation Expansion}
\label{sec:analysis-cooperation}

\begin{figure}[t]
  \centering
  \includegraphics[width=\columnwidth]{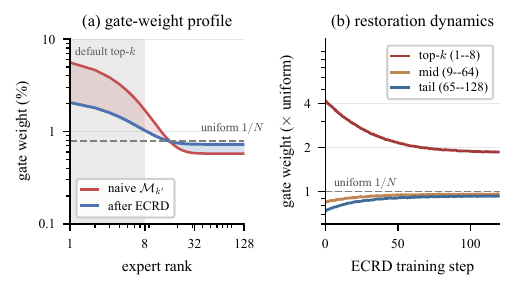}
  \caption{Gate weights on Qwen3-30B-A3B during cooperation restoration. (a) Mean gate weight per expert before and after ECRD, with experts sorted by their base-model weight. The dashed line marks the uniform-weight reference $1/N$. (b) Mean gate weight of the head, mid, and tail groups over ECRD training steps in multiples of $1/N$.}
  \label{fig:gate-qwen}
\end{figure}

To understand what ECRD repairs, Figure~\ref{fig:gate-qwen} examines the gate weights at the router level. An MoE layer outputs $\sum_e g_e\,E_e(x)$, so an expert contributes only in proportion to its gate weight $g_e$, and expanding the budget to $k'{=}N$ activates every expert without changing the router. On Qwen3-30B-A3B the default $k$ experts therefore still carry $26.2\%$ of the gate mass, about $4.2\times$ the uniform-weight reference $1/N$, while the newly activated ones stay below it and contribute little to the layer output. ECRD neutralises this profile by moving weight from the head to the lower-ranked experts, suggesting restored cooperation among the co-activated experts.

\paragraph{Router ablation.} To further examine which components contribute to recovery, we run the ECRD stage on Qwen3-30B-A3B while updating only the router, only the experts, or only the router and experts, and compare them with full parameter ECRD. Training only the router fails to recover the model, with entropy staying extremely low and reward declining. As shown in Table~\ref{tab:router-ablation}, updating only the experts already recovers the model. Jointly updating the experts and the router recovers most of the capability of full parameter ECRD ($90.4\%$). This suggests that recovery relies largely on adapting the co-activated experts. The remaining gap to full parameter ECRD indicates that other components also contribute. Recovery therefore depends not only on reweighting the experts through the router, but also on adapting the experts to their new co-activated partners.

\begin{table}[t]
  \centering
  \small
  \setlength{\tabcolsep}{1.2pt}
  \resizebox{\columnwidth}{!}{%
  \begin{tabular}{lcccccc}
    \toprule
    \textbf{Param} & \textbf{AMC23} & \textbf{Poly-M} & \textbf{AIME24} & \textbf{AIME25} & \textbf{AIME26} & \textbf{Avg.} \\
    \midrule
    All & 71.8 & 48.2 & 27.3 & 20.0 & 20.7 & 37.6 \\
    \midrule
    Routers & \multicolumn{6}{c}{\textit{collapse}} \\
    Experts & 65.0 & 43.8 & 24.3 & 17.7 & 15.7 & 33.3 \\
    R + E & 66.0 & 44.6 & 23.7 & 18.7 & 17.0 & 34.0 \\
    \bottomrule
  \end{tabular}}
  \caption{Mathematics pass@1 (\%) of the activation-expanded Qwen3-30B-A3B after ECRD that updates only the listed components. \emph{All} denotes full parameter ECRD, and \emph{R + E} denotes updating both the routers and the experts. All settings are trained for 12 hours of wall-clock time.}
  \label{tab:router-ablation}
\end{table}

% \begin{figure}[t]
%   \centering
%   \includegraphics[width=0.87\columnwidth]{pics/router_only.pdf}
%   \caption{Training curves of ECRD on the activation-expanded Qwen3-30B-A3B when only the router is trained.}
%   \label{fig:router-only}
% \end{figure}

\subsection{Why the Teacher Must Be Activation-Expanded}
\label{sec:analysis-teacher}

A natural alternative to the activation-expanded teacher $\mathcal{M}_{k'}^{\star}$ is a teacher obtained by applying RL directly to $\mathcal{M}_k$ under the original activation budget. We compare the two from two perspectives: the capability ceiling each teacher can reach, and how much of that capability is transferred to the final low-$k$ model.

\paragraph{The activation-expanded teacher reaches a higher capability ceiling.} We compare the two teachers by their training dynamics and held-out performance. Under the same training budget, CU on the recovered $\mathcal{M}_{k'}$ converges to a higher reward with higher entropy, and outperforms RL on $\mathcal{M}_k$ across all domains and backbones. Detailed comparisons are deferred to Appendix~\ref{sec:appendix-CU-ceiling}.

\paragraph{The stronger teacher is also the better teacher.} A stronger model is not automatically a better teacher for reverse distillation, so we compare the two teacher choices directly. Our results show that the activation-expanded $\mathcal{M}_{k'}^{\star}$ consistently serves as the better teacher, allowing more of the unlocked capability to be transferred back to the original activation budget. Detailed results are deferred to Appendix~\ref{sec:appendix-analysis-teacher-choice}.

\subsection{How Much Capability Survives the Transfer}
\label{sec:analysis-transfer}

Our experiments reveal an interesting phenomenon: the final $k$-activation student $\mathcal{M}_k^{\star}$ can even \emph{surpass} the activation-expanded teacher $\mathcal{M}_{k'}^{\star}$ while still activating only $k$ experts. Detailed comparisons and discussion are deferred to Appendix~\ref{sec:appendix-analysis-student-surpass}.

\section{Conclusion}
\label{sec:conclusion}

In this work, we introduced MoEvolve, a framework that builds a stronger teacher from an MoE model itself without relying on a stronger external teacher or human annotations at the process level. The gains from the expanded model are then distilled back to the original activation budget. Across a wide range of MoE models and benchmarks, our method consistently outperforms the initial models and strong RL baselines at no additional inference cost. Our analysis further shows that the reverse-distilled model can surpass its own activation-expanded teacher. Together, these findings offer a practical recipe for turning the latent capacity of MoE models into real performance gains.

\section*{Limitations}

The proposed MoEvolve framework has not yet been validated on substantially larger MoE models. It has also not been tested on broader tasks such as open-ended generation, agentic tasks, or long-context settings. Whether the observed gains transfer to these regimes remains open. Moreover, constructing a stronger internal teacher relies on training procedures. Thus, such a teacher cannot be obtained in a training-free setting. We currently lack a method to directly unlock the latent expert capacity without additional supervision or optimization.

\section*{Ethical Considerations}

All experiments in this work are conducted on publicly available datasets and open-weight MoE models commonly used in natural language processing research. No human subjects, personal data, or sensitive content are involved. We acknowledge that large language models may carry risks of misuse, bias, or hallucination, and that our framework, which strengthens a model using only its own latent expert capacity, may inherit such behaviors from the underlying backbones.

\bibliography{custom}

\appendix

\section{Model Details}
\label{sec:appendix-models}

Table~\ref{tab:models} summarizes the four MoE backbones used in this work. We use the Instruct version of OLMoE-1B-7B, namely \href{https://huggingface.co/allenai/OLMoE-1B-7B-0924-Instruct}{\nolinkurl{OLMoE-1B-7B-0924-Instruct}} (Apache-2.0 license), and the chat version of DeepSeek-V2-Lite, namely \href{https://huggingface.co/deepseek-ai/DeepSeek-V2-Lite-Chat}{\nolinkurl{DeepSeek-V2-Lite-Chat}} (\href{https://github.com/deepseek-ai/DeepSeek-V2/blob/main/LICENSE-MODEL}{DeepSeek Model License}). For \href{https://huggingface.co/openai/gpt-oss-20b}{GPT-OSS-20B} (Apache-2.0 license), we set the reasoning effort to \texttt{low}, while for \href{https://huggingface.co/Qwen/Qwen3-30B-A3B}{Qwen3-30B-A3B} (Apache-2.0 license), we disable the thinking mode. We use these settings consistently throughout both training and evaluation. All methods share the same backbone, tokenizer, and vocabulary to ensure a fair comparison.

\begin{table}[!hb]
  \centering
  \small
  \setlength{\tabcolsep}{5pt}
  \begin{tabular*}{\columnwidth}{@{\extracolsep{\fill}}lcc@{}}
    \toprule
    \textbf{Model} & \textbf{Params (A/F)} & \textbf{\#Exp.\ (A/F)} \\
    \midrule
    OLMoE-1B-7B & $1.3$/$6.9$B & $8$/$64$ \\
    DeepSeek-V2-Lite & $2.4$/$15.7$B & $6$/$64$ \\
    GPT-OSS-20B & $3.6$/$20.9$B & $4$/$32$ \\
    Qwen3-30B-A3B & $3.3$/$30.5$B & $8$/$128$ \\
    \bottomrule
  \end{tabular*}
  \caption{Overview of the MoE models used in this work. A and F refer to ``Activated'' and ``Full'', and Params are in billions.}
  \label{tab:models}
\end{table}

\section{Evaluation Details}
\label{sec:appendix-eval}

\subsection{Datasets}

All evaluation datasets use English for natural-language content.

\subsubsection{Scientific and General Knowledge (ScienceQA)}

\paragraph{GPQA-main} It contains multiple choice questions written by experts at the graduate level in biology, physics, and chemistry (CC BY 4.0 license) \citep{rein2024gpqa}. We use the main subset to evaluate scientific knowledge and reasoning.

\paragraph{ARC-Challenge} It is the challenging subset of the AI2 Reasoning Challenge, containing multiple-choice science questions that require knowledge and reasoning beyond simple retrieval (CC BY-SA 4.0 license) \citep{clark2018think}.

\paragraph{MMLU-STEM} It comprises the STEM subjects of MMLU, evaluating knowledge through multiple-choice questions across scientific and technical disciplines (MIT license) \citep{hendrycks2021measuring}. We use the test questions from these subjects.

\subsubsection{Mathematics}

\paragraph{AMC23} It contains problems from the 2023 American Mathematics Competitions, covering high school mathematics and reasoning in mathematical competitions \citep{amc2023}. We use the dataset for research evaluation, subject to MAA's copyright and applicable usage requirements.\footnote{MAA's official AMC website: \url{https://maa.org/student-programs/amc/}.}

\paragraph{GSM8K} It consists of grade-school mathematics word problems requiring multi-step arithmetic reasoning (MIT license) \citep{cobbe2021training}. We use its test split to evaluate OLMoE-1B-7B and DeepSeek-V2-Lite.

\paragraph{MATH500} It is an evaluation subset of MATH containing 500 problems, spanning subjects such as algebra, geometry, and number theory (MIT license) \citep{hendrycks2021math}.

\paragraph{Minerva Math} It contains undergraduate mathematics and science problems collected from MIT OpenCourseWare, with numerical or symbolic solutions that can be automatically verified (CC BY-NC-SA 4.0 license for the source course materials) \citep{lewkowycz2022solving}.

\paragraph{PolyMath} It is a multilingual mathematical reasoning benchmark organized into four difficulty levels (Apache-2.0 license for the official repository) \citep{polymath}. We use only its English problems, selecting PolyMath-Low for OLMoE-1B-7B and DeepSeek-V2-Lite and PolyMath-Medium for GPT-OSS-20B and Qwen3-30B-A3B.

\paragraph{AIME} We use problems from the 2024, 2025, and 2026 American Invitational Mathematics Examinations (Apache-2.0) \citep{aime2024,aime2025,aime2026}.

\subsubsection{Code Generation}

\paragraph{HumanEval} It evaluates Python function generation from natural-language docstrings and function signatures (MIT license) \citep{chen2021evaluating}.

\paragraph{MBPP and MBPP-Sanitized} MBPP contains basic Python programming tasks with descriptions in natural language and unit tests (CC BY 4.0 license) \citep{austin2021program}. We evaluate both the full and sanitized versions, with the latter providing a manually reviewed subset with improved task descriptions.

\paragraph{LiveCodeBench-v6} LiveCodeBench collects recent programming competition problems from LeetCode, AtCoder, and Codeforces to evaluate code generation while reducing contamination risk (Creative Commons license; specific variant and version unspecified) \citep{jain2024livecodebench}. We use the lite version of the code generation benchmark (v6), which prunes and samples test cases to reduce dataset size while aiming to preserve evaluation performance.

\subsection{Decoding and Scoring}

All benchmarks are zero-shot at temperature $1.0$, top-$p$ $1.0$, and a maximum generation length of $11{,}000$ tokens. We sample $10$ generations per problem and report the average pass@1 accuracy of each model on each benchmark. We adopt Math-Verify\footnote{\url{https://github.com/huggingface/Math-Verify}} as a rule-based verifier to validate answer correctness on the mathematics benchmarks, where a response is correct when its extracted final answer exactly matches the reference and equivalent surface forms (\textit{e.g.}, different \LaTeX{} spellings of the same expression, or a fraction and its decimal) are accepted. On the multiple-choice ScienceQA benchmarks the extracted option letter has to match the reference exactly, and on the code generation benchmarks a response counts as correct only if it passes \emph{all} unit tests of the problem.

Except for the study in Appendix~\ref{sec:exp-numexperts}, all experiments of \textbf{Ours} use the maximum feasible intermediate budget $k'$.

\section{Implementation Details}
\label{sec:appendix-impl}

\subsection{Training Datasets}
\label{sec:appendix-datasets}

We train on three domains, matching the three evaluation domains. For mathematical reasoning, we mix $25$K problems from \href{https://huggingface.co/datasets/nvidia/OpenMathInstruct-2}{\nolinkurl{OpenMathInstruct-2}} \citep{openmathinstruct2} (CC BY 4.0 license), synthesized from the GSM8K and MATH training sets and therefore easy to medium grade-school and high-school exercises, with $25$K samples from \href{https://huggingface.co/datasets/zwhe99/DeepMath-103K}{\nolinkurl{DeepMath-103K}} \citep{deepmath} (MIT license), drawn from a broader pool of competition and university-level material, for a total of $50$K training instances. Mixing the two difficulty levels keeps the binary reward informative on the weak, just-expanded model while leaving headroom for the stronger ones. For code generation, we use the code split of \href{https://huggingface.co/datasets/PRIME-RL/Eurus-2-RL-Data}{\nolinkurl{Eurus-2-RL-Data}} \citep{eurusrlcode} (MIT license), which consists of $25$K samples. For scientific and general knowledge, we randomly sample $50$K instances from the \href{https://huggingface.co/datasets/cais/mmlu/viewer/auxiliary_train}{\nolinkurl{MMLU-auxiliary_training_set}} \citep{hendrycks2021measuring} (MIT license), which is disjoint from the MMLU-STEM test split used for evaluation. The same data drives all the steps, so all methods share one training corpus. To rule out evaluation contamination, we checked all three corpora against the held-out benchmarks. OpenMathInstruct-2 is synthesized from the GSM8K and MATH \emph{training} splits and is therefore disjoint from their test sets by construction. We additionally verified that no problem has a near-duplicate (5-gram Jaccard $\geq 0.8$) in the AMC23 or AIME-2024/2025/2026 test sets. For Eurus-2-RL-Data, we checked its problem statements against LiveCodeBench-v6 with the same near-duplicate criterion. The MMLU auxiliary training set is a held-out split that is explicitly disjoint from the MMLU-STEM test set used for ScienceQA evaluation. No overlap was found after these checks. All training datasets use English for natural-language content.

\subsection{Training Configuration}
\label{sec:appendix-training}

Table~\ref{tab:hparams} lists the main training configuration used across all methods. Both the OPD objective (Eq.~\eqref{eq:opd}) and the RL step run on the verl framework \citep{sheng2024hybridflow} (Apache-2.0 license). For each backbone and method, we train a separate checkpoint for each domain using only its corresponding training data, and evaluate it only on benchmarks from that domain. To ensure a fair training budget, all methods on the same backbone are trained for the same wall-clock time per domain. Since the backbones differ in training speed, this budget is 8 hours for OLMoE-1B-7B, 16 hours for DeepSeek-V2-Lite, 32 hours for GPT-OSS-20B, and 48 hours for Qwen3-30B-A3B. For MoEvolve, this budget covers both Forward Distillation and Backward Distillation. For all trainable methods, we report the checkpoint with the best validation-set performance.

The RL reward is binary, assigning $1.0$ for a verified-correct math answer, all-passing code unit tests, or a correct multiple-choice option, and $0.0$ otherwise. For both OPD and RL, the batch size of each policy update equals the training batch size, so every batch of rollouts is used for exactly one gradient update. The responses are therefore always sampled from the current policy, which keeps training strictly on-policy.

The OPSD baseline turns $\mathcal{M}_k$ into its own teacher by prompting it with privileged information that is unavailable to the student. Following the original OPSD setup \citep{opsd}, we use the gold option for ScienceQA and the reference solution for mathematics. The code generation training set ships no reference solutions, so we follow SDPO \citep{sdpo} and instead use compiler feedback together with a correct solution from the same rollout group when one is available.

\begin{table}[t]
  \centering
  \resizebox{\columnwidth}{!}{%
  \begin{tabular}{@{}lll@{}}
    \toprule
    \textbf{Group} & \textbf{Setting} & \textbf{Value} \\
    \midrule
    \multirow{5}{*}{\textit{Optimization}}
                                             & Learning rate & $1\times10^{-6}$ \\
                                             & Warm-up steps & 0 \\
                                             & Optimizer & Adam \\
    \midrule
    \multirow{2}{*}{\textit{Sequence}} & Maximum prompt length & 2000 \\
                                         & Maximum response length & 11000 \\
    \midrule
    \multirow{3}{*}{\textit{Rollout}} & Rollout $n$ & 8 \\
                                        & Sampling temperature & 1.0 \\
                                        & Top-$p$ & 1.0 \\
    \midrule
    \multirow{6}{*}{\textit{RL algorithm}} & Advantage estimator & SAPO \\
                                             & Use KL in reward & False \\
                                             & KL loss coefficient & 0.0 \\
                                             & Entropy coefficient & 0.0 \\
                                             & $\tau_{\mathrm{pos}}$ & 1.0 \\
                                             & $\tau_{\mathrm{neg}}$ & 1.05 \\
    \midrule
    \multirow{4}{*}{\textit{OPD algorithm}} & loss\_mode & k1 \\
                                              & use\_policy\_gradient & true \\
                                              & distillation\_loss\_coef & 1.0 \\
                                              & ExOPD reward scale $\lambda$ & 1.25 \\
    \midrule
    \multirow{4}{*}{\textit{ECRD stopping}} & Phase-switch EMA $\alpha$ & 0.9 \\
                                              & Stopping threshold $\rho$ & 0.01 \\
                                              & Convergence window $W$ & 20 steps \\
                                              & Convergence patience $P$ & 3 windows \\
    \midrule
    \textit{Hardware} & GPUs & 16 $\times$ NVIDIA H20 \\
    \bottomrule
  \end{tabular}%
  }
  \caption{Main training configuration.}
  \label{tab:hparams}
\end{table}

\subsection{EMA-Based Stopping Criterion for ECRD}
\label{sec:appendix-ema-analyses}

Since the sole purpose of the ECRD step is to re-establish expert cooperation, we do not fix the number of distillation steps in advance. Instead, we switch to capability recovery when the distillation objective stops making meaningful progress. To avoid reacting to noisy single-step fluctuations, we first maintain an exponential moving average (EMA) of the absolute distillation objective. Let $a_s = |\mathcal{J}_{\text{OPD}}^{(s)}|$ at step $s$. We set $e_1=a_1$ and update
\begin{equation}
    e_s = \alpha e_{s-1} + (1-\alpha)a_s,
    \label{eq:stage1-ema}
\end{equation}
where $\alpha$ is the EMA coefficient. We then perform convergence checks every $W$ steps. If $b$ is the beginning of the current window and $s=b+W$, we compute the relative EMA drop
\begin{equation}
    r_s = \frac{e_b - e_s}{\max(e_b, \epsilon)}.
    \label{eq:stage1-rel-drop}
\end{equation}
A window is marked as converged when $r_s < \rho$, where $\rho$ is a minimum relative-drop threshold. Otherwise, the convergence counter is reset. After each check, the window baseline is rolled forward to the current $(e_s,s)$. Once $P$ consecutive windows are marked as converged, we regard the distillation objective as having reached a plateau and transition to capability recovery. The complete ECRD procedure is summarized in Algorithm~\ref{alg:method}.

% Figure~\ref{fig:opd-stop} shows that this rule fires at the right moment. The distillation objective falls from $0.40$ to about $0.045$, but it looks flat long before cooperation is actually restored, so simpler heuristics stop far too early. A fixed budget of $50$ steps, an absolute threshold of objective $<0.1$, and a short window without patience ($W{=}5$, $P{=}1$) stop at steps $50$, $66$, and $101$, where the reward is still only $0.36$, $0.42$, and $0.53$ and the gate weights of the co-activated experts are still moving by several percent per window. Our rule stops at step $221$, where the reward has plateaued at the pre-RL level of the teacher $\mathcal{M}_k$ and the gate-weight drift has fallen below $0.3\%$ per window. The two signals we care about, task reward and expert weights, therefore agree that the expanded experts have returned to a cooperative regime, so the model is ready for capability recovery. Removing the patience requirement ($P{=}1$) stops at step $181$, which is already close but rests on a single window, so we keep $P{=}3$ to guard against one spuriously flat window.

\begin{algorithm}[t]
\caption{Expert Cooperation Restoration Distillation}
\label{alg:method}
\begin{algorithmic}[1]
\Require MoE $\mathcal{M}_k$, target budget $k'>k$, prompts $\mathcal{D}$, EMA coefficient $\alpha$, window size $W$, threshold $\rho$, patience $P$
\Ensure Recovered activation-expanded model $\mathcal{M}_{k'}$
\State Initialize $\mathcal{M}_{k'}$ by changing only the activation budget from $k$ to $k'$
\State $e \gets \textsc{None}$; $b \gets 1$; $c \gets 0$ \Comment{EMA, window start, converged-window counter}
\For{distillation step $s=1,2,\ldots$}
    \State Sample trajectories from student $\mathcal{M}_{k'}$ on prompts from $\mathcal{D}$
    \State Update $\mathcal{M}_{k'}$ with teacher $\mathcal{M}_k$ using OPD
    \State $a_s \gets |\mathcal{L}_{\mathrm{OPD}}^{(s)}|$
    \State $e \gets a_s$ if $e$ is \textsc{None}, otherwise $e \gets \alpha e + (1-\alpha)a_s$
    \If{$s-b = W$}
        \State $r \gets (e_b-e)/\max(e_b,\epsilon)$
        \State $c \gets c+1$ if $r<\rho$, otherwise $c\gets 0$
        \State $e_b \gets e$; $b \gets s$
        \If{$c=P$} \textbf{break} \EndIf
    \EndIf
\EndFor
\State \Return recovered $\mathcal{M}_{k'}$
\end{algorithmic}
\end{algorithm}

\section{Intermediate Activation Width under Larger Activation Budgets}
\label{sec:appendix-capacity}

This section formalizes the intuition used in \S\ref{sec:method}, namely that raising the activation budget from $k$ to $k'$ widens the active intermediate representation of every MoE layer, exposing more pretrained parameters to each token. We follow the notation of \S\ref{sec:prelim}, where $\mathbf{h} \in \mathbb{R}^d$ is the hidden representation, $\mathcal{T}_k(\mathbf{h})$ is the set of top-$k$ experts selected by the router, and $\tilde{g}_i(\mathbf{h})$ are the normalized routing weights obtained from Eq.~\eqref{eq:router}. Modern MoE LLMs instantiate each expert $E_i$ as a gated feed-forward network. Writing $d_{\mathrm{e}}$ for the intermediate width of a single expert,
\begin{equation}
    E_i(\mathbf{h}) = W_{2,i}\big(\sigma(W_{g,i}\mathbf{h}) \odot W_{u,i}\mathbf{h}\big),
    \label{eq:expert-ffn}
\end{equation}
where $W_{g,i}, W_{u,i} \in \mathbb{R}^{d_{\mathrm{e}} \times d}$, $W_{2,i} \in \mathbb{R}^{d \times d_{\mathrm{e}}}$, $\sigma$ is an elementwise activation, and $\odot$ is the elementwise product. We write $\mathbf{z}_i(\mathbf{h}) = \sigma(W_{g,i}\mathbf{h}) \odot W_{u,i}\mathbf{h} \in \mathbb{R}^{d_{\mathrm{e}}}$ for the intermediate activation of expert $i$. To make the activation budget explicit we also write $\mathbf{y}_k(\mathbf{h})$ for the layer output $\mathbf{y}$ of \S\ref{sec:prelim} under budget $k$. Substituting Eq.~\eqref{eq:expert-ffn} into that output gives
\begin{equation}
    \mathbf{y}_k(\mathbf{h}) = \!\!\sum_{i \in \mathcal{T}_k(\mathbf{h})}\!\! \tilde{g}_i(\mathbf{h})\, W_{2,i}\, \mathbf{z}_i(\mathbf{h}),
    \label{eq:moe-sum}
\end{equation}
which is a sum of matrix-vector products. Since $\tilde{g}_i(\mathbf{h})$ is a scalar, it can be absorbed into the output projection as $\widehat{W}_{2,i}(\mathbf{h}) = \tilde{g}_i(\mathbf{h}) W_{2,i}$, which turns Eq.~\eqref{eq:moe-sum} into
\begin{equation}
    \mathbf{y}_k(\mathbf{h}) = \!\!\sum_{i \in \mathcal{T}_k(\mathbf{h})}\!\! \widehat{W}_{2,i}(\mathbf{h})\, \mathbf{z}_i(\mathbf{h}).
    \label{eq:moe-absorbed}
\end{equation}
Let $\mathcal{T}_k(\mathbf{h}) = \{i_1,\dots,i_k\}$. Block matrix multiplication states that for any $A_j \in \mathbb{R}^{d \times d_{\mathrm{e}}}$ and $v_j \in \mathbb{R}^{d_{\mathrm{e}}}$,
\begin{equation}
    \sum_{j=1}^{k} A_j v_j = \big[A_1 \cdots A_k\big]\big[v_1^\top \cdots v_k^\top\big]^\top,
    \label{eq:block-identity}
\end{equation}
so the concatenated form below is an identity rather than an additional assumption. Applying Eq.~\eqref{eq:block-identity} to Eq.~\eqref{eq:moe-absorbed} with $A_j = \widehat{W}_{2,i_j}(\mathbf{h})$ and $v_j = \mathbf{z}_{i_j}(\mathbf{h})$ yields
\begin{equation}
\begin{aligned}
    \mathbf{y}_k(\mathbf{h}) &= \mathbf{W}_2^{\mathcal{T}_k}(\mathbf{h})\, \mathbf{z}_{\mathcal{T}_k}(\mathbf{h}), \quad \text{with} \\
    \mathbf{W}_2^{\mathcal{T}_k}(\mathbf{h}) &= \big[\widehat{W}_{2,i_1}(\mathbf{h}) \cdots \widehat{W}_{2,i_k}(\mathbf{h})\big], \\
    \mathbf{z}_{\mathcal{T}_k}(\mathbf{h}) &= \big[\mathbf{z}_{i_1}(\mathbf{h})^\top \cdots \mathbf{z}_{i_k}(\mathbf{h})^\top\big]^\top,
\end{aligned}
    \label{eq:moe-blocked}
\end{equation}
where $\mathbf{W}_2^{\mathcal{T}_k}(\mathbf{h}) \in \mathbb{R}^{d \times k d_{\mathrm{e}}}$ and $\mathbf{z}_{\mathcal{T}_k}(\mathbf{h}) \in \mathbb{R}^{k d_{\mathrm{e}}}$. It remains to check that $\mathbf{z}_{\mathcal{T}_k}(\mathbf{h})$ is itself a gated feed-forward activation of width $k d_{\mathrm{e}}$. Because $\sigma$ and $\odot$ act elementwise, they commute with vertical concatenation, so stacking the per-expert projections into $\mathbf{W}_g^{\mathcal{T}_k}, \mathbf{W}_u^{\mathcal{T}_k} \in \mathbb{R}^{k d_{\mathrm{e}} \times d}$ gives
\begin{equation}
    \mathbf{z}_{\mathcal{T}_k}(\mathbf{h}) = \sigma\big(\mathbf{W}_g^{\mathcal{T}_k}\mathbf{h}\big) \odot \big(\mathbf{W}_u^{\mathcal{T}_k}\mathbf{h}\big).
    \label{eq:wide-intermediate}
\end{equation}
Combining Eqs.~\eqref{eq:moe-blocked} and~\eqref{eq:wide-intermediate} gives the explicit form
\begin{equation}
    \mathbf{y}_k(\mathbf{h}) = \mathbf{W}_2^{\mathcal{T}_k}(\mathbf{h})\big(\sigma(\mathbf{W}_g^{\mathcal{T}_k}\mathbf{h}) \odot \mathbf{W}_u^{\mathcal{T}_k}\mathbf{h}\big),
    \label{eq:wide-ffn}
\end{equation}
so the top-$k$ MoE layer is exactly a gated feed-forward layer whose intermediate width is $k d_{\mathrm{e}}$, with a block-structured and routing-dependent output projection. Eq.~\eqref{eq:wide-intermediate} therefore makes the role of the activation budget explicit. Since $\mathcal{M}_k$ and $\mathcal{M}_{k'}$ share the same expert and routing parameters and differ only in how many experts are activated, moving from $k$ to $k'$ widens the active intermediate representation from $\mathbb{R}^{k d_{\mathrm{e}}}$ to $\mathbb{R}^{k' d_{\mathrm{e}}}$ and adds $(k'-k)\,d_{\mathrm{e}}$ active intermediate features per token without introducing any new parameter. Activation expansion thus exposes additional capacity that is already encoded in the pretrained weights but left unused under the original budget, which ECRD and CU are designed to exploit.

\section{Training Instability without Expert Cooperation Restoration}
\label{sec:appendix-reward-entropy}

\begin{figure}[t]
  \centering
  \includegraphics[width=0.48\textwidth]{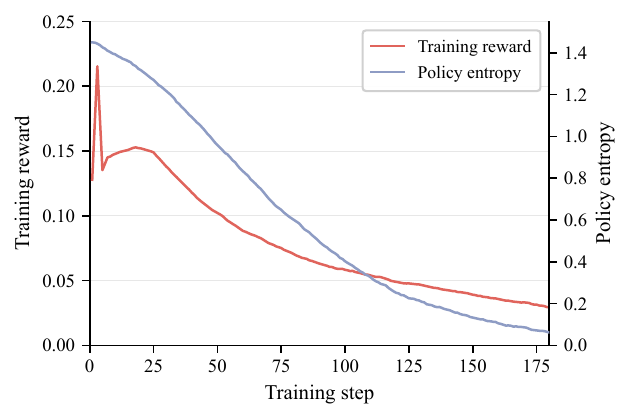}
  \caption{Training curves of applying RL directly on the activation-expanded DeepSeek-V2-Lite, without prior cooperation restoration.}
  \label{fig:reward-entropy}
\end{figure}

As shown in Figure~\ref{fig:reward-entropy}, when RL is applied directly to the activation-expanded DeepSeek-V2-Lite without the ECRD restoration step, training quickly becomes unstable. The policy entropy collapses early, while the training reward fails to improve consistently, which supports our decision to restore expert cooperation before applying RL in the CU stage.

\begin{figure}[t]
  \centering
  \includegraphics[width=\columnwidth]{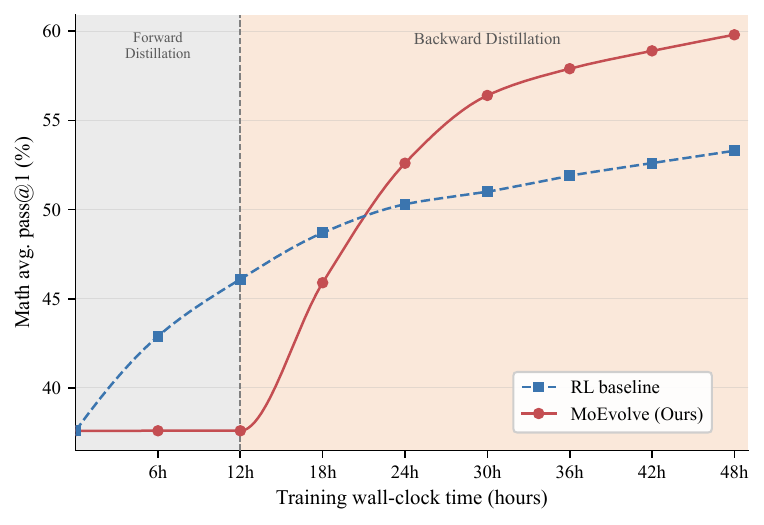}
  \caption{Math avg.\ pass@1 \textit{vs.}\ wall-clock training time on Qwen3-30B-A3B. The grey region is Forward Distillation, during which $\mathcal{M}_k$ is unchanged, and the orange region is Backward Distillation, where capability is transferred back to $\mathcal{M}_k$. The dashed blue line shows the RL baseline trained directly on $\mathcal{M}_k$ for the same total time.}
  \label{fig:compute}
\end{figure}

\section{Computational Cost}
\label{sec:appendix-compute}

Figure~\ref{fig:compute} plots the mathematics average pass@1 against wall-clock training time on Qwen3-30B-A3B. During Forward Distillation the original $\mathcal{M}_k$ is unchanged, so its score stays at the base level. For each point in the Backward Distillation region, the time remaining after Forward Distillation is split between CU and RD, with $80\%$ used for CU and $20\%$ for RD, and we evaluate the model obtained after RD. For example, the point at 24 hours uses 9.6 hours for CU and 2.4 hours for RD. Once Backward Distillation begins, the capability transferred from the stronger teacher drives a rapid improvement. The RL baseline is trained directly on $\mathcal{M}_k$ for the same total wall-clock time. MoEvolve reaches a higher final score despite spending the first 12 hours of wall-clock time building the internal teacher.

% \begin{table*}[t]
%   \centering
%   \footnotesize
%   \setlength{\tabcolsep}{4pt}
%   \begin{tabular*}{\textwidth}{@{\extracolsep{\fill}}l ccc ccc ccc ccc@{}}
%     \toprule
%     \multirow{2}{*}{\textbf{Objective}} & \multicolumn{3}{c}{\textbf{OLMoE-1B-7B}} & \multicolumn{3}{c}{\textbf{DeepSeek-V2-Lite}} & \multicolumn{3}{c}{\textbf{GPT-OSS-20B}} & \multicolumn{3}{c}{\textbf{Qwen3-30B-A3B}} \\
%     \cmidrule(lr){2-4} \cmidrule(lr){5-7} \cmidrule(lr){8-10} \cmidrule(lr){11-13}
%     & \textbf{Sci.} & \textbf{Math} & \textbf{Code} & \textbf{Sci.} & \textbf{Math} & \textbf{Code} & \textbf{Sci.} & \textbf{Math} & \textbf{Code} & \textbf{Sci.} & \textbf{Math} & \textbf{Code} \\
%     \midrule
%     RFT & 44.6 & 15.9 & 20.4 & 53.2 & 34.4 & 34.0 & 74.3 & 55.9 & 69.0 & 81.2 & 56.5 & 68.4 \\
%     OPD & 44.7 & 16.1 & 20.6 & 53.2 & 34.5 & 34.3 & 75.3 & 56.7 & 69.6 & 81.4 & 57.5 & 68.6 \\
%     \textbf{ExOPD} & \textbf{45.7} & \textbf{17.7} & \textbf{22.6} & \textbf{53.8} & \textbf{36.2} & \textbf{35.8} & \textbf{75.4} & \textbf{59.3} & \textbf{70.6} & \textbf{82.0} & \textbf{59.8} & \textbf{69.1} \\
%     \bottomrule
%   \end{tabular*}
%   \caption{Ablation on the reverse-distillation objective (Eq.~\eqref{eq:gopd}). Given the same recovered teacher $\mathcal{M}_{k'}^{\star}$, the original $k$-activation model $\mathcal{M}_k$ is distilled back using RFT, vanilla OPD, or ExOPD. We report the average score within each domain (Sci.\ denotes ScienceQA).}
%   \label{tab:reverse-ds}
% \end{table*}

\section{Analysis of Training Dynamics and Expert Load}
\label{sec:appendix-analysis-dynamics}

\begin{figure*}[tbp]
  \centering
  \includegraphics[width=0.95\textwidth]{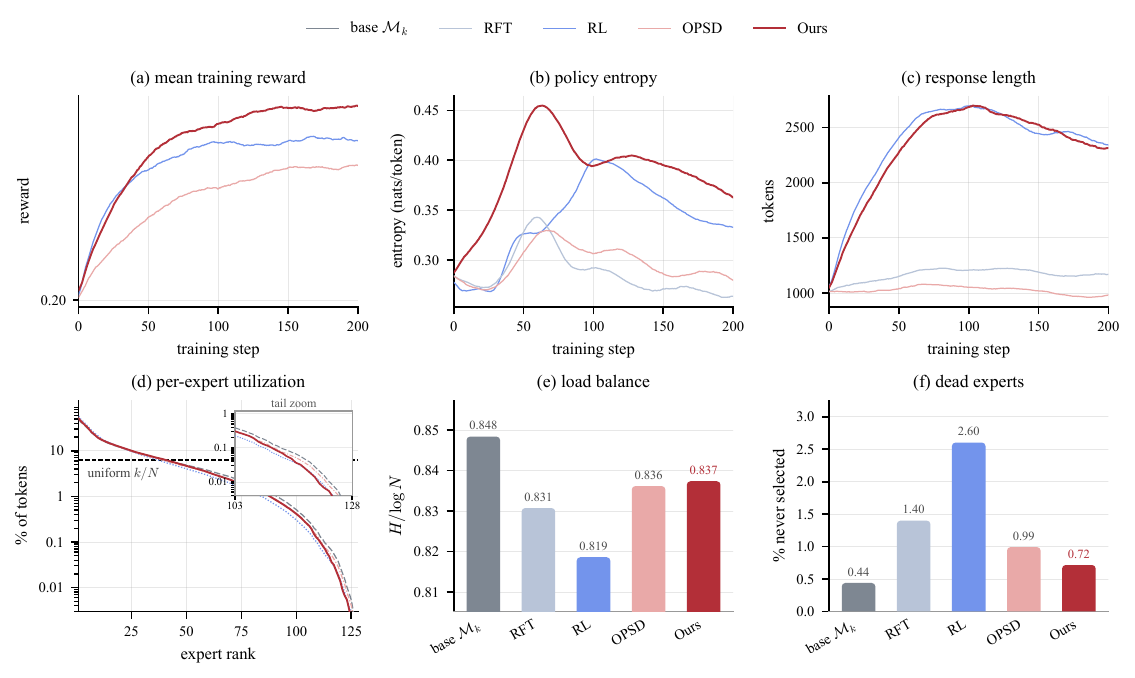}
  \caption{Training dynamics (top) and final expert load (bottom) on Qwen3-30B-A3B. The top row shows mean reward, policy entropy, and response length over training. The bottom row shows per-expert utilization, load balance $H/\log N$, and the fraction of never-selected expert slots.}
  \label{fig:dynamics-load}
\end{figure*}

We compare the methods both during training and in the routing patterns of the resulting models on Qwen3-30B-A3B ($k{=}8$, $k'{=}128$, $N{=}128$ experts in each of its $48$ MoE layers). All results in Figure~\ref{fig:dynamics-load} are from the mathematics domain. All methods start from the same $k$-activation checkpoint (Figure~\ref{fig:dynamics-load}(a)--(c)). For MoEvolve, the curves are taken from the RD stage, where $\mathcal{M}_k$ is trained. RL raises reward fastest but saturates early. Its policy entropy keeps climbing during the first half of training and then declines. RFT and OPSD show smaller entropy rises than RL. RFT falls back the fastest and ends below its initial level, while OPSD returns to its initial level. Both converge to lower reward. MoEvolve improves for longer, finishes with the highest reward, and retains the highest entropy, suggesting that it remains exploratory at convergence. Response length rises for RL and MoEvolve in the first half of training and then falls back, whereas RFT and OPSD stay close to the initial length, with RFT slightly longer and OPSD ending slightly shorter. The routing statistics of the final $k$-activation models tell the same story (Figure~\ref{fig:dynamics-load}(d)--(f)). With $128$ experts per layer, the informative part of the load profile is the starved tail. Every trained checkpoint concentrates routing relative to the base model $\mathcal{M}_k$, but the fixed-budget baselines do so much more strongly, pushing traffic to the hottest experts, reducing load balance $H/\log N$, and leaving several times more experts never selected, with RL the most skewed. Our model remains closest to the routing of $\mathcal{M}_k$ ($H/\log N$ of $0.837$ vs.\ $0.848$, and $0.72\%$ vs.\ $0.44\%$ of the $6144$ expert slots never selected), reaching the highest reward while preserving almost all inherited expert capacity.

\begin{figure}[tbp]
  \centering
  \includegraphics[width=\columnwidth]{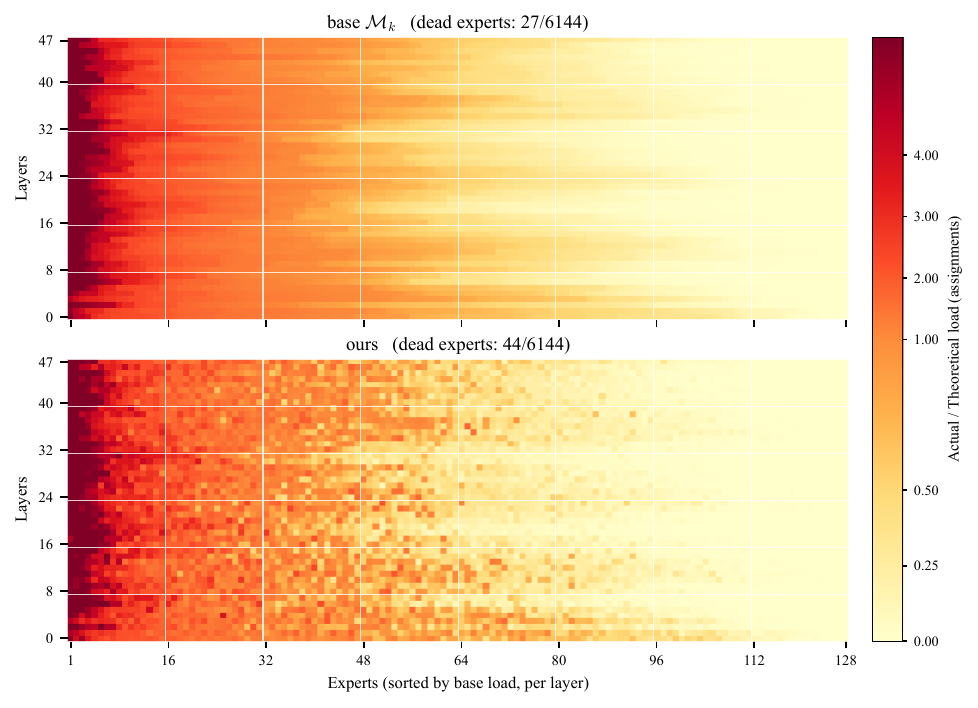}
  \caption{Per-layer, per-expert load of Qwen3-30B-A3B ($k{=}8$, $N{=}128$, $48$ MoE layers) before (top) and after (bottom) our training, measured on the mathematics benchmarks. A cell is the actual/theoretical load of a (layer, expert) slot, so $1.0$ (white) is the uniform-load reference $k/N$, red is overloaded, pale is starved, and $0$ means the expert is never selected. The number of such dead slots is given in each panel title. Experts are sorted per layer by their load in the base model and the same permutation is used in both panels.}
  \label{fig:heatmap-qwen}
\end{figure}

\begin{table*}[t]
  \centering
  \footnotesize
  \setlength{\tabcolsep}{4pt}
  \begin{tabular*}{\textwidth}{@{\extracolsep{\fill}}l ccc ccc ccc ccc@{}}
    \toprule
    \multirow{2}{*}{\textbf{Objective}} & \multicolumn{3}{c}{\textbf{OLMoE-1B-7B}} & \multicolumn{3}{c}{\textbf{DeepSeek-V2-Lite}} & \multicolumn{3}{c}{\textbf{GPT-OSS-20B}} & \multicolumn{3}{c}{\textbf{Qwen3-30B-A3B}} \\
    \cmidrule(lr){2-4} \cmidrule(lr){5-7} \cmidrule(lr){8-10} \cmidrule(lr){11-13}
    & \textbf{Sci.} & \textbf{Math} & \textbf{Code} & \textbf{Sci.} & \textbf{Math} & \textbf{Code} & \textbf{Sci.} & \textbf{Math} & \textbf{Code} & \textbf{Sci.} & \textbf{Math} & \textbf{Code} \\
    \midrule
    RFT & 44.6 & 15.9 & 20.4 & 53.2 & 34.4 & 34.0 & 74.3 & 55.9 & 69.0 & 81.2 & 56.5 & 68.4 \\
    OPD & 44.7 & 16.1 & 20.6 & 53.2 & 34.5 & 34.3 & 75.3 & 56.7 & 69.6 & 81.4 & 57.5 & 68.6 \\
    \textbf{ExOPD} & \textbf{45.7} & \textbf{17.7} & \textbf{22.6} & \textbf{53.8} & \textbf{36.2} & \textbf{35.8} & \textbf{75.4} & \textbf{59.3} & \textbf{70.6} & \textbf{82.0} & \textbf{59.8} & \textbf{69.1} \\
    \bottomrule
  \end{tabular*}
  \caption{Ablation on the reverse-distillation objective (Eq.~\eqref{eq:gopd}). Given the same recovered teacher $\mathcal{M}_{k'}^{\star}$, the original $k$-activation model $\mathcal{M}_k$ is distilled back using RFT, vanilla OPD, or ExOPD. We report the average score within each domain (Sci.\ denotes ScienceQA).}
  \label{tab:reverse-ds}
\end{table*}

\paragraph{Expert-Load Heat Maps Before and After Our Training}
\label{sec:appendix-heatmap}

Figure~\ref{fig:heatmap-qwen} resolves the routing statistics of Figure~\ref{fig:dynamics-load}(d)--(f) per expert. It compares the base model $\mathcal{M}_k$ with our final $k$-activation model $\mathcal{M}_k^{\star}$ on Qwen3-30B-A3B, aggregated over the same mathematics benchmark decoding traces. The two maps are almost indistinguishable. Our training only slightly darkens the head and pales the tail of each layer. It adds $17$ dead slots out of $6144$ (from $0.44\%$ to $0.72\%$), and even lowers the hottest slot from $11.8\times$ to $10.7\times$ the uniform-load reference. The gains of Section~\ref{sec:analysis} are therefore obtained without reorganising which experts carry the traffic.

% \begin{figure*}[tbp]
%   \centering
%   \includegraphics[width=0.85\textwidth]{pics/moe_heatmap_qwen.pdf}
%   \caption{Per-layer, per-expert load of Qwen3-30B-A3B ($k{=}8$, $N{=}128$, $48$ MoE layers) before (top) and after (bottom) our training, measured on the mathematics benchmarks. A cell is the actual/theoretical load of a (layer, expert) slot, so $1.0$ (white) is the uniform-load reference $k/N$, red is overloaded, pale is starved, and $0$ means the expert is never selected. The number of such dead slots is given in each panel title. Experts are sorted per layer by their load in the base model and the same permutation is used in both panels.}
%   \label{fig:heatmap-qwen}
% \end{figure*}

\section{Ablation on the Reverse-Distillation Objective}
\label{sec:appendix-rd-ablation}

We ablate the reverse-distillation objective in Eq.~\eqref{eq:gopd}, where the recovered activation-expanded model $\mathcal{M}_{k'}^{\star}$ is the teacher and the original $k$-activation model $\mathcal{M}_k$ is the student. We compare three ways to transfer the teacher's capability back into the original activation budget. \textbf{RFT} imitates the teacher's own correct trajectories off-policy. \textbf{OPD} instead supervises the student's own trajectories with the token-level KL objective in Eq.~\eqref{eq:opd}, on-policy but still bounded by the teacher. \textbf{ExOPD} adds reward extrapolation \citep{yang2026learning}, letting the student go beyond the teacher.

As shown in Table~\ref{tab:reverse-ds}, the ranking ExOPD\,$>$\,OPD\,$>$\,RFT is consistent across backbones. RFT$\to$OPD gains about $+0.1$ to $+0.8$ average points, while adding reward extrapolation yields a further $+1.1$ to $+1.5$ points. This is expected because the activation-expanded teacher $\mathcal{M}_{k'}^{\star}$ is stronger than the original-budget student, leaving $\mathcal{M}_k$ more room to explore and improve, while reward extrapolation turns that teacher--student gap into a cleaner token-level learning signal than plain OPD or the off-policy targets used by RFT.

\section{Experiments on the Effect of the Number of Activated Experts}
\label{sec:exp-numexperts}

\begin{figure}[t]
  \centering
  \includegraphics[width=\columnwidth]{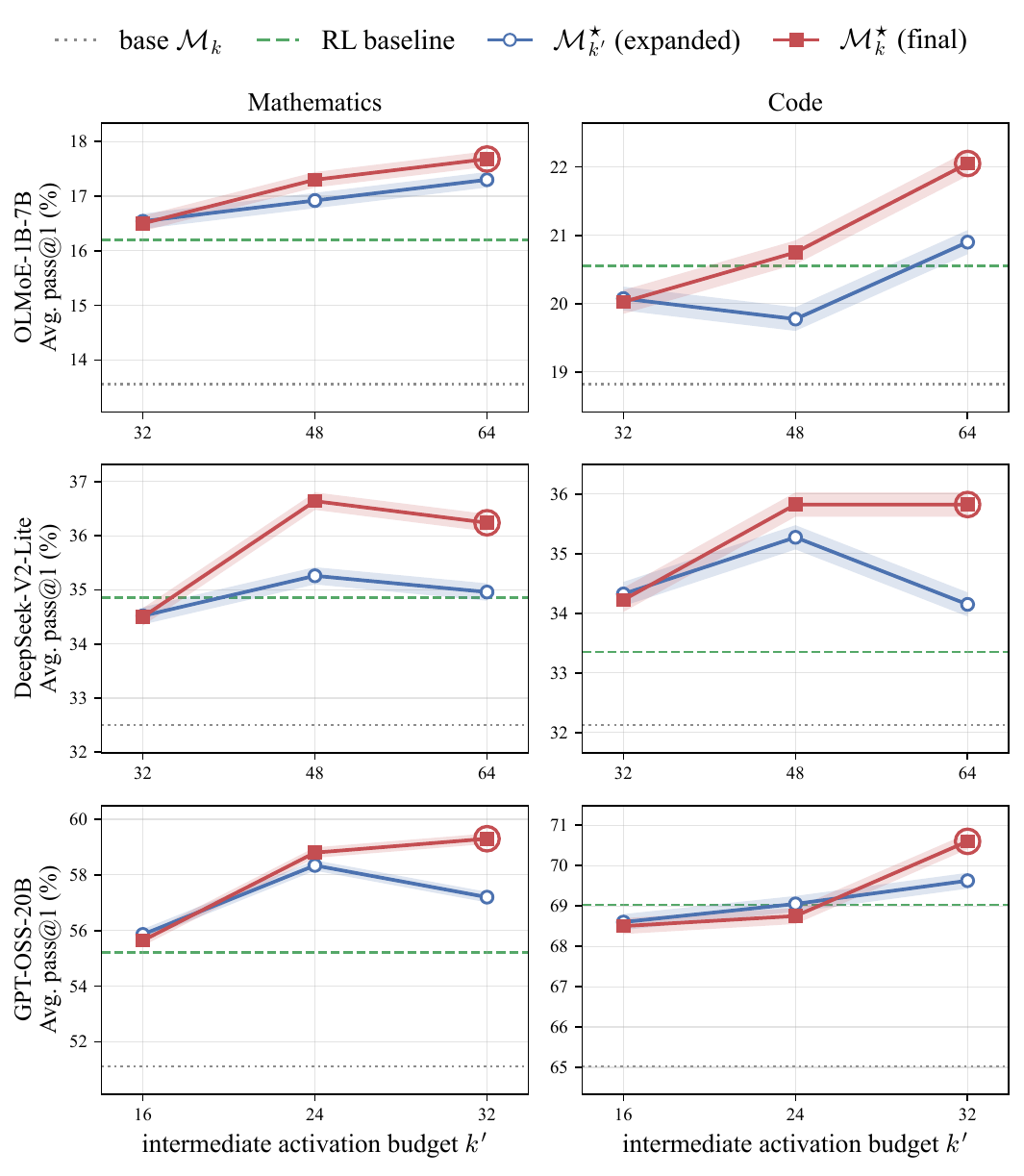}
  \caption{Effect of the intermediate activation budget $k'$ on the three backbones. Error bars indicate the standard error across 3 seeds.}
  \label{fig:numexperts}
\end{figure}

We vary the activation budget $k'$ of the activation-expanded model to study how the size of the intermediate model affects the final reverse-distilled model. Figure~\ref{fig:numexperts} summarizes, for the three backbones on which we can afford the full sweep, the mathematics and code averages of both the recovered activation-expanded teacher $\mathcal{M}_{k'}^{\star}$ and the final model $\mathcal{M}_k^{\star}$ obtained after reverse distillation (always $k$ experts), as a function of $k'$. The original base model and the fixed-budget RL baseline are drawn as horizontal references. We use the maximum feasible activation budget as a simple default rather than selecting $k'$ on the test set. The sweep shows that this choice is consistently strong: OLMoE-1B-7B and GPT-OSS-20B peak at the maximum budgets $k'{=}64$ and $k'{=}32$, while DeepSeek-V2-Lite remains close to its peak when using the maximum budget $k'{=}64$. This means that our method does not require careful tuning of $k'$ and that the maximal feasible activation budget already provides a simple and effective default.

% \begin{figure*}[t]
%   \centering
%   \includegraphics[width=0.98\textwidth]{pics/numexperts.pdf}
%   \caption{Effect of the intermediate activation budget $k'$ on the three backbones. The top row shows mathematics average pass@1, and the bottom row shows code average pass@1. Blue plots the recovered activation-expanded teacher $\mathcal{M}_{k'}^{\star}$, red plots the final reverse-distilled model $\mathcal{M}_k^{\star}$, and the circled marker marks the maximum feasible budget used in the main experiments. Dashed green and dotted grey lines denote the RL on $\mathcal{M}_k$ baseline and the original model $\mathcal{M}_k$. Error bars indicate the standard error across 3 seeds.}
%   \label{fig:numexperts}
% \end{figure*}

% \section{Analyses}
% \label{sec:appendix-analysis}

\section{Why CU unlocks a higher capability ceiling}
\label{sec:appendix-CU-ceiling}
\begin{figure}[tbp]
  \centering
  \includegraphics[width=0.9\columnwidth]{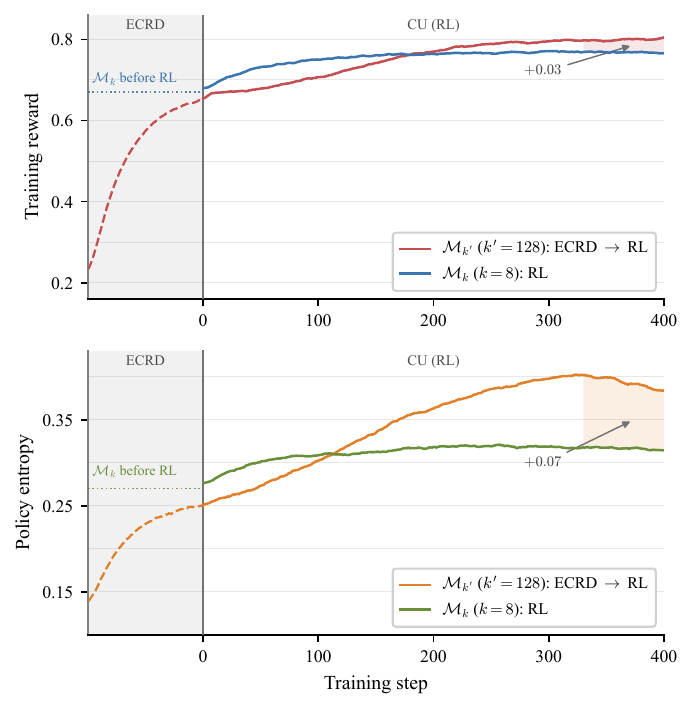}
  \caption{Comparisonof training reward (top) and policy entropy (bottom) on Qwen3-30B-A3B between RL on recovered $\mathcal{M}_{k'}$ and RL on $\mathcal{M}_k$.}
  \label{fig:teacher-reward}
\end{figure}

\begin{figure}[tbp]
  \centering
  \includegraphics[width=\columnwidth]{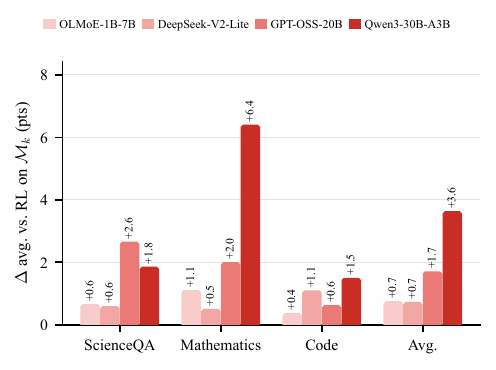}
  \caption{Held-out advantage of the activation-expanded teacher $\mathcal{M}_{k'}^{\star}$ over RL on the original $\mathcal{M}_k$, in pass@1 points. Bars report the three domain averages.}
  \label{fig:teacher-eval}
\end{figure}

Figure~\ref{fig:teacher-reward} shows the training dynamics on Qwen3-30B-A3B. During ECRD, the activation-expanded model climbs from the severe degradation caused by naive expansion back to just below the reward level of $\mathcal{M}_k$, indicating that this stage mainly restores a trainable cooperative regime. Once CU begins, however, $\mathcal{M}_{k'}$ continues improving for much longer than RL on $\mathcal{M}_k$, eventually overtakes it, and converges about $0.03$ reward higher. Its policy entropy also stays higher throughout CU, suggesting broader exploration and a less collapsed policy. The same ordering appears in held-out evaluation, where Figure~\ref{fig:teacher-eval} shows that $\mathcal{M}_{k'}^{\star}$ beats RL on $\mathcal{M}_k$ in every domain of every backbone.

\section{Analysis of Teacher Choice}
\label{sec:appendix-analysis-teacher-choice}

Since a stronger model is not automatically a better teacher for reverse distillation, we fix the student to $\mathcal{M}_k$ and the reverse-distillation objective to ExOPD in both cases, and only change the teacher. As shown in Table~\ref{tab:teacher-choice}, using RL on $\mathcal{M}_k$ as the teacher improves over the original model only modestly, while replacing it with $\mathcal{M}_{k'}^{\star}$ yields a clearly stronger student on every backbone. This shows that the activation-expanded $\mathcal{M}_{k'}^{\star}$ is not only a stronger model, but also a better teacher for reverse distillation, allowing more of the unlocked capability to be transferred back to the original activation budget.

\begin{table}[t]
  \centering
  \footnotesize
  \setlength{\tabcolsep}{3.5pt}
  \begin{tabular*}{\columnwidth}{@{\extracolsep{\fill}}lccc@{}}
    \toprule
    \multirow{2}{*}{\textbf{Backbone}} & \multirow{2}{*}{\textbf{$\mathcal{M}_k$}} & \multicolumn{2}{c}{\textbf{ExOPD-teacher}} \\
    \cmidrule(lr){3-4}
    & & \textbf{RL on $\mathcal{M}_k$} & \textbf{$\mathcal{M}_{k'}^{\star}$} \\
    \midrule
    OLMoE-1B-7B & 13.6 & 16.3 & \textbf{17.7} \\
    DeepSeek-V2-Lite & 32.5 & 33.9 & \textbf{36.2} \\
    GPT-OSS-20B & 51.1 & 55.7 & \textbf{59.3} \\
    Qwen3-30B-A3B & 37.6 & 53.1 & \textbf{59.8} \\
    \bottomrule
  \end{tabular*}
  \caption{Teacher-choice comparison for RD on the mathematics average.}
  \label{tab:teacher-choice}
\end{table}

\section{Analysis of Student Surpassing the Teacher}
\label{sec:appendix-analysis-student-surpass}

Table~\ref{tab:student-surpass} shows that the student surpasses the teacher in most settings. On OLMoE-1B-7B, DeepSeek-V2-Lite, and GPT-OSS-20B, the student is ahead on both the mathematics and code averages, for example $36.2$ \textit{vs.}\ $35.3$ on mathematics for DeepSeek-V2-Lite and $59.3$ \textit{vs.}\ $57.2$ for GPT-OSS-20B. Qwen3-30B-A3B is the exception, where the two are on par on mathematics and the teacher remains slightly ahead on code. A plausible explanation, consistent with the reward-extrapolation view of ExOPD \citep{yang2026learning}, is that the RD step does not merely match the teacher. Instead, the stronger teacher $\mathcal{M}_{k'}^{\star}$ leaves more room for the student $\mathcal{M}_k$ to explore and improve, allowing it to go beyond simple teacher matching relative to the reference model while adapting the transferred behavior to its own $k$-expert capacity. This suggests that the internally constructed teacher provides well-matched supervision that can support capability gains beyond teacher imitation.

\begin{table}[t]
  \centering
  \small
  \setlength{\tabcolsep}{2.8pt}
  \begin{tabular*}{\columnwidth}{@{\extracolsep{\fill}}lcccc@{}}
    \toprule
    \multirow{2}{*}{\textbf{Backbone}} & \multicolumn{2}{c}{\textbf{Math avg.}} & \multicolumn{2}{c}{\textbf{Code avg.}} \\
    \cmidrule(lr){2-3} \cmidrule(lr){4-5}
    & \textbf{$\mathcal{M}_{k'}^{\star}$} & \textbf{$\mathcal{M}_k^{\star}$} & \textbf{$\mathcal{M}_{k'}^{\star}$} & \textbf{$\mathcal{M}_k^{\star}$} \\
    \midrule
    OLMoE-1B-7B & 17.3 & \textbf{17.7} & 20.9 & \textbf{22.6} \\
    DeepSeek-V2-Lite     & 35.3 & \textbf{36.2} & 34.4 & \textbf{35.8} \\
    GPT-OSS-20B               & 57.2 & \textbf{59.3} & 69.6 & \textbf{70.6} \\
    Qwen3-30B-A3B             & 59.7 & \textbf{59.8} & \textbf{69.5} & 69.1 \\
    \bottomrule
  \end{tabular*}
  \caption{Comparison between the activation-expanded teacher $\mathcal{M}_{k'}^{\star}$ and the final reverse-distilled student $\mathcal{M}_k^{\star}$ on mathematics and code averages.}
  \label{tab:student-surpass}
\end{table}

\section{Prompt Templates}
\label{sec:appendix-prompts}

\begin{figure}[!htb]
  \centering
  \includegraphics[width=\columnwidth]{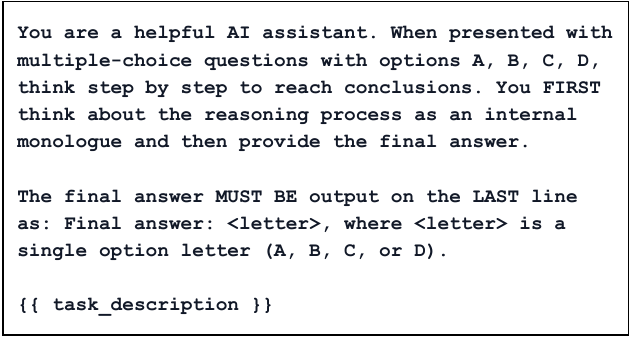}
  \caption{ScienceQA prompt template.}
  \label{fig:prompt-qa}
\end{figure}

\begin{figure}[!htb]
  \centering
  \includegraphics[width=\columnwidth]{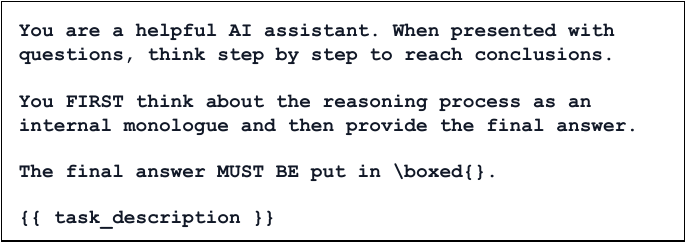}
  \caption{Mathematics prompt template.}
  \label{fig:prompt-math}
\end{figure}

\begin{figure}[!htb]
  \centering
  \includegraphics[width=\columnwidth]{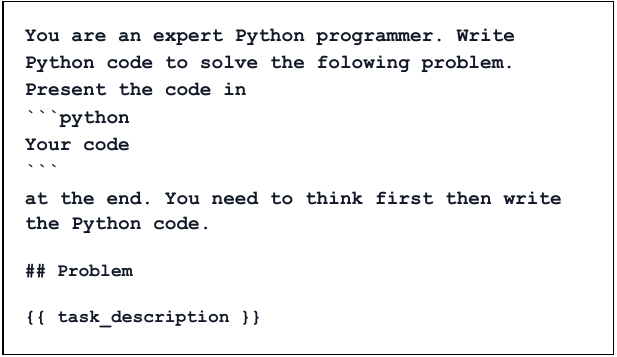}
  \caption{Code prompt template.}
  \label{fig:prompt-code}
\end{figure}

The prompt templates used throughout training and evaluation for the ScienceQA, mathematics, and code domains are shown in Figures~\ref{fig:prompt-qa}, \ref{fig:prompt-math}, and \ref{fig:prompt-code}, respectively.

\section{AI Use Statement}
\label{app:ai_use_statement}
We used ChatGPT and Claude to assist in drafting and refining the text, solely to improve clarity, readability, and style through suggestions on wording, phrasing, and grammar. All AI-assisted content was reviewed, revised, and verified by the authors. These tools did not play an independent role in research ideation, experimental design, data collection, data analysis, or interpretation of results. The authors made all substantive intellectual contributions and bear full responsibility for all content in this manuscript.

% \begin{figure}[!htb]
%   \centering
%   \includegraphics[width=\columnwidth]{pics/codesys.drawio.pdf}
%   \caption{Code prompt template.}
%   \label{fig:prompt-code}
% \end{figure}

% \suppressfloats[t]
% \begin{figure}[!t]
%   \centering
%   \captionsetup{skip=6pt, belowskip=55pt}
%   \includegraphics[width=\columnwidth]{pics/mathsys.drawio.pdf}
%   \caption{Mathematics prompt template.}
%   \label{fig:prompt-math}

%   \includegraphics[width=\columnwidth]{pics/codesys.drawio.pdf}
%   \caption{Code prompt template.}
%   \label{fig:prompt-code}

%   \includegraphics[width=\columnwidth]{pics/qasys.drawio.pdf}
%   \caption{ScienceQA prompt template.}
%   \label{fig:prompt-qa}
% \end{figure}

\end{document}